\documentclass{article} 
\usepackage[preprint]{colm2026_conference} 

\usepackage{microtype}
\usepackage{graphicx}
\usepackage{subcaption}
\usepackage{booktabs}

\usepackage{hyperref}

\usepackage{url}
\usepackage{enumitem}

\usepackage{lineno}

\definecolor{darkblue}{rgb}{0, 0, 0.5}
\hypersetup{colorlinks=true, citecolor=darkblue, linkcolor=darkblue, urlcolor=darkblue}

\usepackage{amsmath}
\usepackage{amssymb}
\usepackage{mathtools}
\usepackage{amsthm}

\usepackage[capitalize,noabbrev]{cleveref}

\theoremstyle{plain}

\theoremstyle{definition}

\theoremstyle{remark}

\usepackage[textsize=tiny]{todonotes}

\usepackage[utf8]{inputenc} 
\usepackage[T1]{fontenc}    
\usepackage{amsmath}
\usepackage{amssymb}
\usepackage{amsthm}
\usepackage{multirow}        
\usepackage{tcolorbox}

\usepackage[draft]{minted} 
\setminted{
  breaksymbolleft={},
  breaksymbolright={}
}
\usepackage{newunicodechar}

\newunicodechar{ℕ}{\ensuremath{\mathbb{N}}}
\newunicodechar{ℚ}{\ensuremath{\mathbb{Q}}}
\newunicodechar{ℤ}{\ensuremath{\mathbb{Z}}}
\newunicodechar{ℝ}{\ensuremath{\mathbb{R}}}
\newunicodechar{ℂ}{\ensuremath{\mathbb{C}}}

\newunicodechar{↑}{\ensuremath{\uparrow}}
\newunicodechar{↓}{\ensuremath{\downarrow}}
\newunicodechar{→}{\ensuremath{\rightarrow}}
\newunicodechar{←}{\ensuremath{\leftarrow}}
\newunicodechar{⟶}{\ensuremath{\longrightarrow}}
\newunicodechar{⇒}{\ensuremath{\Rightarrow}}
\newunicodechar{⇐}{\ensuremath{\Leftarrow}}
\newunicodechar{⇔}{\ensuremath{\Leftrightarrow}}

\newunicodechar{₀}{\ensuremath{_0}}
\newunicodechar{₁}{\ensuremath{_1}}
\newunicodechar{₂}{\ensuremath{_2}}
\newunicodechar{₃}{\ensuremath{_3}}
\newunicodechar{₄}{\ensuremath{_4}}
\newunicodechar{₅}{\ensuremath{_5}}
\newunicodechar{₆}{\ensuremath{_6}}
\newunicodechar{₇}{\ensuremath{_7}}
\newunicodechar{₈}{\ensuremath{_8}}
\newunicodechar{₉}{\ensuremath{_9}}
\newunicodechar{∣}{\ensuremath{\mid}}
\newunicodechar{≤}{\ensuremath{\leq}}
\newunicodechar{≥}{\ensuremath{\geq}}
\newunicodechar{≠}{\ensuremath{\neq}}
\newunicodechar{≡}{\ensuremath{\equiv}}
\newunicodechar{≈}{\ensuremath{\approx}}

\newunicodechar{∃}{\ensuremath{\exists}}
\newunicodechar{∀}{\ensuremath{\forall}}
\newunicodechar{∧}{\ensuremath{\land}}
\newunicodechar{∨}{\ensuremath{\lor}}
\newunicodechar{¬}{\ensuremath{\lnot}}
\newunicodechar{⊥}{\ensuremath{\bot}}
\newunicodechar{⊤}{\ensuremath{\top}}

\newunicodechar{α}{\ensuremath{\alpha}}
\newunicodechar{β}{\ensuremath{\beta}}
\newunicodechar{γ}{\ensuremath{\gamma}}
\newunicodechar{δ}{\ensuremath{\delta}}
\newunicodechar{ε}{\ensuremath{\varepsilon}}
\newunicodechar{λ}{\ensuremath{\lambda}}
\newunicodechar{μ}{\ensuremath{\mu}}
\newunicodechar{π}{\ensuremath{\pi}}
\newunicodechar{σ}{\ensuremath{\sigma}}
\newunicodechar{τ}{\ensuremath{\tau}}
\newunicodechar{φ}{\ensuremath{\varphi}}
\newunicodechar{ω}{\ensuremath{\omega}}

\newunicodechar{∈}{\ensuremath{\in}}
\newunicodechar{∉}{\ensuremath{\notin}}
\newunicodechar{⊆}{\ensuremath{\subseteq}}
\newunicodechar{⊇}{\ensuremath{\supseteq}}
\newunicodechar{∪}{\ensuremath{\cup}}
\newunicodechar{∩}{\ensuremath{\cap}}
\newunicodechar{∅}{\ensuremath{\emptyset}}

\newunicodechar{⟦}{\ensuremath{\llbracket}}
\newunicodechar{⟧}{\ensuremath{\rrbracket}}

\newunicodechar{⊢}{\ensuremath{\vdash}}
\newunicodechar{⊣}{\ensuremath{\dashv}}

\newunicodechar{∑}{\ensuremath{\Sigma}}
\newunicodechar{↔}{\ensuremath{\leftrightarrow}}
\newunicodechar{Γ}{\ensuremath{\Gamma}}
\newunicodechar{Ω}{\ensuremath{\Omega}}
\newunicodechar{─}{\ensuremath{\textemdash}}
\newunicodechar{ᵀ}{\ensuremath{\textsuperscript{T}}}
\newunicodechar{ᵥ}{\ensuremath{_v}}
\newunicodechar{⬝}{\ensuremath{\cdot}}
\newunicodechar{ₗ}{\ensuremath{_l}}
\newunicodechar{≪}{\ensuremath{\ll}}
\newunicodechar{≫}{\ensuremath{\gg}}
\newunicodechar{ᶜ}{\ensuremath{\textsuperscript{c}}}
\newunicodechar{ˣ}{\ensuremath{\textsuperscript{x}}}
\newunicodechar{∠}{\ensuremath{\angle}}
\newunicodechar{∟}{\ensuremath{\angle}}
\newunicodechar{⟫}{\ensuremath{\rangle}}
\newunicodechar{⟪}{\ensuremath{\langle}}
\newunicodechar{∂}{\ensuremath{\partial}}
\newunicodechar{∫}{\ensuremath{\int}}
\newunicodechar{ᵐ}{\ensuremath{\textsuperscript{m}}}

\def\framework{DreamProver}

\makeatletter
\newcommand{\titlenote}[1]{%
  \begingroup
  \renewcommand\thefootnote{}%
  \thanks{#1}%
  \addtocounter{footnote}{-1}%
  \endgroup
}
\makeatother

\newcommand{\smallsec}[1]{\textbf{#1.}}

\begin{document}


\title{\framework{}: Evolving Transferable Lemma Libraries via a Wake-Sleep Theorem-Proving Agent}

\author{
\parbox{0.9\linewidth}{
Youyuan Zhang$^{1*}$,
Jialiang Sun$^{1*}$,
Hangrui Bi$^{1}$,
Chuqin Geng$^{1}$,
Wenjie Ma$^{2}$,
Zhaoyu Li$^{1\dagger}$,
Xujie Si$^{1\dagger}$
}
\\
$^{1}$University of Toronto \qquad $^{2}$UC Berkeley
\titlenote{Equal contribution. $\dagger$Equal advising.}
}
\ifcolmsubmission
\linenumbers
\fi

\maketitle



\begin{abstract}
We introduce \framework{}, an agentic framework that leverages a ``wake-sleep'' program induction paradigm to discover reusable lemmas for formal theorem proving. Existing approaches either rely on fixed lemma libraries, which limit adaptability, or synthesize highly specific intermediate lemmas tailored to individual theorems, thereby lacking generality. 
\framework{} addresses this gap through an iterative two-stage process. In the \emph{wake} stage, \framework{} attempts to prove theorems from a training set using the current lemma library while proposing new candidate lemmas. In the \emph{sleep} stage, it abstracts, refines, and consolidates these candidates to compress and optimize the library. Through this alternating cycle, \framework{} progressively evolves a compact set of high-level, transferable lemmas that can be effectively used to prove unseen theorems in related domains.
Experimental results demonstrate that \framework{} substantially improves proof success rates across a diverse set of mathematical benchmarks, while also producing more concise proofs and reducing computational cost.
\end{abstract}
\section{Introduction}
\label{introduction}
Automated theorem proving has recently seen rapid progress~\citep{li2024survey,yang2024formal,weng2025autoformalization}, driven by advances in large language models (LLMs) and the growing ecosystem of formal proof assistants such as Lean~\citep{lean}. On one hand, LLMs provide powerful reasoning capabilities for planning and generating proofs. On the other hand, large-scale formal libraries (e.g., mathlib~\citep{mathlib}), developed by the active community, offer rich repositories of mathematical definitions and theorems across diverse domains. Together, these developments have substantially expanded the scope and accessibility of automated theorem proving. As a result, state-of-the-art systems can now solve increasingly complex mathematical problems, achieving performance that rivals or even surpasses human experts on high-school and undergraduate-level competitions~\citep{deepseekproverv2,chen2025seed,achim2025aristotle}, while making promising strides toward research-level challenges~\citep{gao2026archon,angdinata2026paucity}.

Despite these advances, most existing approaches treat each theorem largely in isolation, without explicitly leveraging past experience to improve future proving~\citep{lego}. In particular, while proofs often draw on lemmas from existing libraries in the training data or are decomposed into multiple intermediate lemmas, these lemmas are typically constructed in a detailed, problem-specific manner and rarely generalize across theorems~\citep{dsp}. These limitations stand in stark contrast to human mathematical practice, where people systematically accumulate and abstract a core set of transferable lemmas from prior problems, enabling more efficient problem solving within related mathematical domains.

To address this limitation, we present \framework{}, a theorem-proving agent that explicitly learns and consolidates reusable lemmas over time. Inspired by the ``wake-sleep'' algorithm~\citep{ellis2021dreamcoder}, \framework{} operates through an iterative two-stage cycle. In the \emph{wake} stage, the system attempts to solve formal mathematical theorems from a training set, sketching proofs guided by the current lemma library and recursively decomposing problems into smaller sub-problems that can be proved using existing lemmas. In the \emph{sleep} stage, the system analyzes the structural representations of the accumulated lemmas, first clustering and pruning them based on semantic similarity, and then performing abstraction over each cluster to produce a compact and validated lemma library. Crucially, these learned lemmas are carried forward and refined across iterations, enabling \framework{} to progressively evolve transferable, domain-specific knowledge, rather than repeatedly relying on LLMs to generate problem-specific intermediate lemmas or on existing libraries.

Experiments on mathematical benchmarks spanning diverse domains, including inequality~\citep{wei2024proving, lips}, number theory~\citep{putnambench, deepseekproverv2}, combinatorics~\citep{combibench}, plane geometry~\citep{leangeo}, and machine learning theory~\citep{formalml}, demonstrate that \framework{} significantly improves proving success rates over prior state-of-the-art approaches by 61\% using the learned lemma library. In addition, it generates more concise proofs under a smaller compute budget, reducing proof length by 50\% and token usage by 48\%.

\section{Related Work}
\label{related_work}
\subsection{Theorem Proving with LLMs}
LLMs have demonstrated strong capabilities in formal theorem proving~\citep{li2024survey,yang2024formal}, with state-of-the-art systems achieving silver- and gold-medal-level performance on the International Mathematical Olympiad~\citep{alphaproof,chen2025seed,achim2025aristotle}.  A line of work focuses on tactic-based approaches, where LLMs are trained to predict single proof steps and are combined with search strategies to incrementally construct proofs~\citep{bfsv1,bfsv2,li2024hunyuanprover,wu2024internlm2,dong2025stp}. In contrast, Draft-Sketch-Prove~\citep{dsp,dsp_plus} generates high-level proof sketches for whole-proof generation and subsequently verifies intermediate steps. Recent approaches train LLMs for direct whole-proof generation on large-scale formal corpora using supervised fine-tuning and reinforcement learning~\citep{wang2025kimina,deepseekprover,deepseekproverv15,deepseekproverv2}, including DeepSeek-Prover and Goedel-Prover model families~\citep{deepseekproverv2,goedelv2}. 
Beyond these paradigms, another line of work explores recursive subgoal decomposition, which breaks down difficult theorems into simpler subgoals with iterative error correction~\citep{recursive,dong2024formal,delta}, exemplified by Hilbert~\citep{varambally2025hilbertrecursivelybuildingformal}. Still, these approaches largely treat each theorem independently and do not reuse lemmas across problems. 

\subsection{Library Learning}
Library learning aims to accumulate intermediate results that can be reused across problems. It has been widely studied in program synthesis~\citep{ellis2021dreamcoder,topdownsynthesis,lilo}, robotics~\citep{quest,iscil,lotus}, and planning~\citep{parl,lads} as a mechanism for discovering reusable abstractions. In theorem proving, library learning involves maintaining abstractions of existing lemmas and proposing reusable lemmas across problems. Lego-Prover~\citep{lego} introduces a growing lemma library with modular reuse; however, subsequent analysis~\citep{lego_fail} finds no clear evidence of effective reuse or performance gains under controlled computational budgets. Divide and Abstract~\citep{divideandabstract} incorporates abstraction learning into autoformalization. 
Seed-Prover~\citep{chen2025seed} maintains a lemma pool of proved conjectures and selects candidates based on difficulty and semantic relevance, but its library is instance-specific and does not persist across problems. 
In contrast, \framework{} continually abstracts, consolidates, and prunes lemmas into a persistent library of reusable knowledge, enabling improved performance on future theorem proving tasks.

\section{Method}
\label{method}

\begin{figure*}[t]
  \centering
  \vspace{-1em}
  \includegraphics[width=0.9\textwidth]{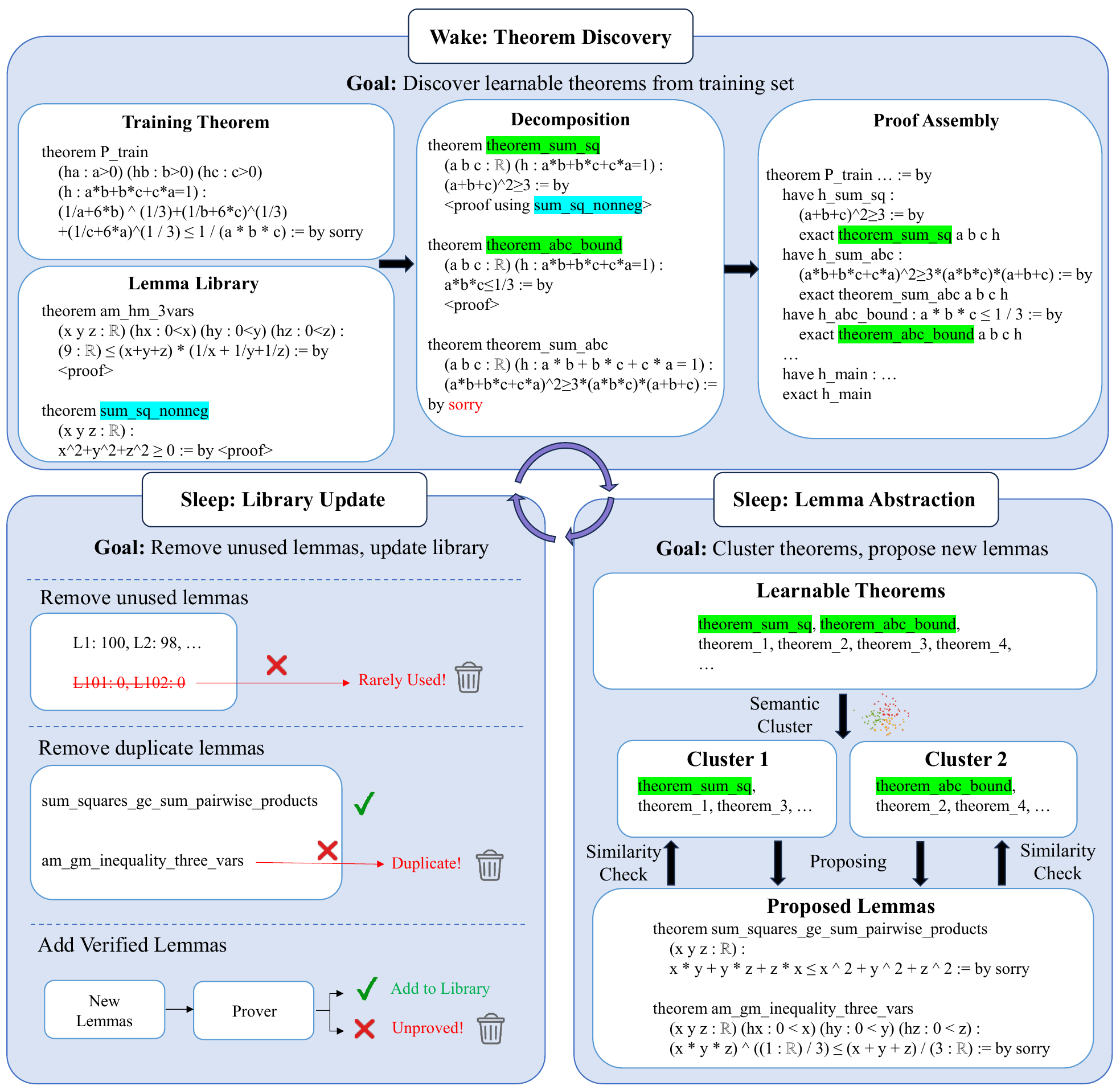}
  \caption{Overview of \framework{} as an iterative wake--sleep framework for learning a reusable lemma library. 
In the wake phase (top), the system attempts to prove a target theorem $x$ using the current library $L$, 
where an LLM decomposes the problem into subgoals and constructs a proof $p_x$ by leveraging existing lemmas. 
In the sleep phase (bottom), the resulting subgoals and intermediate theorems are semantically clustered 
and abstracted to generate candidate lemmas. Redundant, low-quality, or unused lemmas are pruned, and 
verified, high-utility lemmas are incorporated back into $L$, enabling continual refinement of a compact 
and transferable library. }
  \label{fig:wake_sleep_overview}
  \vspace{-1em}
\end{figure*}

Inspired by the wake-sleep algorithm~\citep{hinton1995wake, ellis2021dreamcoder}, \framework{} operates in iterative wake–sleep cycles that mimic the human cognitive process of learning. Throughout these cycles, the system maintains a library that stores important skills, which can be repeatedly reused to solve new problems. In the wake stage, the system explores and acquires new experience by attempting to prove as many theorems as possible from the training set using the knowledge from the current library. The resulting experiences are formulated as intermediate theorems, which are derived during the proving process from the training problems. In the sleep stage, the system learns from these experiences and summarizes them into new skills. Specifically, it first aggregates and abstracts the derived intermediate theorems to evolve more general and reusable lemmas, and then updates the library by optimizing it to maintain both conciseness and reusability. 

In this section, we first introduce the wake stage, which is responsible for intermediate theorem discovery, followed by the sleep stage, which performs lemma evolution and library learning. Finally, we describe how the learned library is used to prove unseen problems in the test set. The overview of \framework{} is shown in Figure~\ref{fig:wake_sleep_overview}.

\subsection{Wake Stage}

The wake stage is designed to enable the system to explore unsolved problems within the domain and acquire new experiences for learning. In the setting of formal theorem proving, the system learns reusable skills in the form of lemmas by attempting to prove theorems from the training set. The exploration begins from scratch, with an empty lemma library. As the process iterates, the library gradually grows: in each wake stage, the system is equipped with the current lemma library and attempts to solve problems using the lemmas it contains. Through this process, the system discovers meaningful learning signals in the form of intermediate theorems, which serve as the basis for inducing new lemmas during the subsequent sleep stage. Next, we elaborate on the two key steps through which the system derives intermediate theorems from the training set.

\smallsec{Learnable theorem identification} While proving theorems in the training set, the system identifies and collects learnable theorems. By design, a theorem is learnable by the system if it can be \emph{directly} proved by the set of currently learned lemmas. To identify whether a target theorem is learnable given the knowledge level entailed by the current lemma library, the system attempts to prove the theorem by directly prompting an LLM. We first fix the compute budget by defining the maximum numbers of proof attempts and error-correction attempts. If the LLM is able to solve the problem using the lemmas from the library under the given compute budget, then the theorem is considered as learnable and potentially implies a more advanced lemma built on the existing lemmas. Such theorems are collected and subject to the evolving process in the sleep stage.

\smallsec{Intermediate theorem discovery} When a theorem in the training set is challenging to prove directly, \framework{} recursively generates intermediate theorems that are sufficiently tractable for the LLM to prove and learn from. This is achieved via recursive theorem decomposition~\citep{recursive,goedelv2,varambally2025hilbertrecursivelybuildingformal}. Specifically, if the system fails to solve a theorem through direct prompting with the current lemma library, it decomposes the target theorem into a set of simpler sub-theorems that collectively contribute to constructing a complete proof.

Given a theorem from the training set, the decomposition process first generates a formal proof sketch, and then extracts intermediate theorems from the unproven parts of the sketch (i.e., intermediate goals marked by the \texttt{sorry} tactic). Both steps are performed via LLM prompting. Notably, extracting intermediate goals as standalone theorems is non-trivial. To ensure their validity, the system attempts to reassemble the full proof by combining the proof sketch with the extracted intermediate theorems and verifying whether the target theorem can be proved. In addition, the system leverages an LLM to check the proof sketch in natural language, enabling efficient filtering of potentially incorrect decompositions.

In practice, during the early iterations when the lemma library is still limited and lacks sufficient abstraction, it is often difficult to prove theorems directly or even with a single level of decomposition. As the library improves over iterations, more theorems from the training set can be solved either directly or with only one level of decomposition.

\subsection{Sleep Stage}

In the sleep stage, the system learns from past experiences by summarizing them into transferable skills. In the context of theorem proving, this stage is responsible for evolving new lemmas and optimizing the lemma library. To achieve this, two main steps are executed in sequence. First, the system performs lemma abstraction over the collected learnable lemmas to derive more general and reusable forms. Second, the system incorporates the proposed new lemmas into the library, updating it to improve both its quality and utility.

\smallsec{Lemma abstraction} To extract common and useful knowledge from the intermediate theorems collected during the wake stage, \framework{} adopts a cluster-based lemma evolution strategy. In general, we first group the theorems based on their semantic meaning, and then propose new candidate lemmas for each cluster. The candidate lemmas are further validated by measuring their formal structural similarity to the theorems within the cluster.

Specifically, we first annotate each theorem with a natural language description. The description provides high-level explanations of the theorem, including its sub-domain and difficulty (e.g., whether it involves a trivial equivalence transformation or requires more advanced reasoning). In addition, the description includes relevant theorems that can be used in its proof. We generate these annotations by prompting an LLM to reason about the theorem and produce the corresponding descriptions. The prompt template used for this process is provided in Appendix~\ref{app:prompts}.

Next, we compute embeddings of the descriptions and perform clustering in the semantic embedding space. The embeddings are obtained using a sentence transformer~\citep{song2020mpnet}, and distances are measured using cosine similarity between embedding vectors. We then apply the K-Means algorithm for clustering, where the number of clusters is determined using the elbow method~\citep{elbow}.

\framework{} then performs abstraction over each cluster, where the grouped theorems and their corresponding descriptions are provided as input context to an LLM to generate candidate lemmas that can be used to prove the theorems in the cluster. To further validate that the candidate lemmas proposed from this coarse abstraction are applicable, we measure the structural similarity~\citep{tbps} between each candidate lemma and the theorems in the cluster, and retain a lemma if its maximum similarity exceeds a predefined threshold. Specifically, to compute structural similarity, we first convert both the candidate lemma and each theorem into simplified first-order logic expression trees, and then measure similarity between these tree representations. The similarity score reflects how well a source expression tree can be aligned with a target expression tree, indicating whether the proposed lemma is potentially useful for proving the theorems in the cluster.

Through this process, the experiences collected during the wake stage are consolidated into transferable skills in the form of newly evolved lemmas, which are then incorporated into the lemma library for future use.

\smallsec{Library update} To enable \framework{} to effectively reuse evolved lemmas during theorem proving, \framework{} stores newly acquired skills by updating the lemma library. This update consists of three main steps: (1) forgetting previously unused lemmas, (2) storing new lemmas, and (3) formally verifying the new lemmas.

As an essential mechanism in the learning process, forgetting potentially non-reusable skills makes room for more important and frequently used ones. \framework{} implements this mechanism during the sleep stage by removing less reusable lemmas. Specifically, \framework{} adopts a least-recently-used strategy that tracks the usage frequency of each lemma across past wake stages, and removes those that are least frequently used once the library reaches its maximum capacity. Empirically, this forgetting mechanism maintains a concise lemma library of fewer than 100 lemmas, enabling stable and efficient reuse. It also avoids the unexpected exclusion of important lemmas when the entire library is directly incorporated as input context.

To store new lemmas, \framework{} adds non-duplicate ones to the existing lemma library. To determine whether a newly proposed lemma already exists in the library, \framework{} leverages tree edit distance, which measures the cost of transforming one expression tree into another, to compute similarity between the candidate lemma and each existing lemma. Lemmas with high structural similarity (i.e., low edit distance) are considered duplicates and are removed from the candidate list. This step reduces redundancy and preserves the diversity of the lemma library.

Finally, each remaining candidate lemma is formally verified by attempting to prove it using the same direct prompting strategy employed in the wake stage. Notably, the proposed lemmas are typically well-known theorems with clean and concise forms (see Appendix~\ref{app:lemma_library}), making them relatively easy to prove under a small sampling budget.

\subsection{Inference}

\framework{} adopts a lightweight inference framework. It first attempts to prove the theorem via direct prompting with the lemma library. If that attempt fails, it falls back to a sketch-and-prove workflow: it first generates a proof sketch, then proves each subgoal individually, again using the lemma library during both sketch generation and subgoal solving. By leveraging the library throughout the process, the system can prove unseen test problems more effectively while using fewer tokens.
\section{Experiments}
\label{experiments}

Since lemma libraries are inherently domain-specific, we evaluate \framework{} across multiple mathematical domains. Specifically, we consider both domains well represented in LLM training data (e.g., mathlib~\citep{mathlib}) and domains underrepresented in such data. Our experiments aim to answer the following research questions:

\begin{enumerate}[label=\textbf{RQ\arabic*:}]
    \item \textbf{Well-Represented Domain Capability.} Can \framework{} prove more theorems than existing methods in domains well captured by LLM knowledge?
    
    \item \textbf{Inference Efficiency and Proof Quality.} Can \framework{} operate under smaller inference budgets while producing more concise and higher-quality proofs?
    
    \item \textbf{Effectiveness of Learned Lemmas.} Does \framework{} effectively reuse evolved lemmas when solving unseen problems?
    
    \item \textbf{Underrepresented Domain Generalization.} Can \framework{} prove more theorems than existing methods in domains underrepresented in LLM knowledge?
\end{enumerate}

\subsection{RQ1: Well-Represented Domain Capability}
\label{exp:main_results}

\smallsec{Datasets} We evaluate performance on well-represented domains across three mathematical areas: number theory, inequalities, and combinatorics.
For inequalities, we use three established benchmarks from existing formalizations~\citep{lips}: 567NEQ~\citep{Tung2012Nice}, ChenNEQ~\citep{chen2014brief}, and MO-INT~\citep{wei2024proving}. For number theory, we consider the subsets of PutnamBench~\citep{putnambench} and ProverBench~\citep{deepseekproverv2} restricted to this domain. For combinatorics, we use a subset of CombiBench~\citep{combibench}, where each problem is paired with its provided solution as the target theorem.
To learn the lemma library, we construct domain-specific libraries using separate training sets. For inequalities, we sample 100 problems from the AIPS~\citep{wei2024proving} training set. For number theory and combinatorics, we similarly sample 100 problems from the corresponding training subsets of FormalMATH~\citep{formalmath}.

\smallsec{Baselines} We compare \framework{} with three categories of LLM-based baselines: proprietary reasoning models, open-source prover models, and agentic theorem-proving systems. For proprietary LLMs, we report pass@32 results under one-shot proof generation for GPT-5.3-Codex, Claude Opus 4.6, Gemini 2.5 Pro, and Gemini 3.1 Pro Preview. For open-source prover LLMs, we evaluate DeepSeek-Prover-V2-7B~\citep{deepseekproverv2} and Goedel-Prover-V2-8B/32B~\citep{lin2025goedel}. Following the Goedel-Prover-V2 setup, we apply self-correction with pass@32 and allow up to three refinement iterations. For agentic systems, we use Hilbert~\citep{varambally2025hilbertrecursivelybuildingformal} with a maximum decomposition depth of 2 to control the sampling budget, while \framework{} is limited to a single decomposition step. Implementation details for both Hilbert and \framework{} are provided in Appendix~\ref{app:implementation_details}.

\begin{table*}[ht!]
\centering
\resizebox{\textwidth}{!}{
\begin{tabular}{l ccc cc c}
\toprule
\multirow{2}{*}{\textbf{Method}} &
\multicolumn{3}{c}{\textbf{Inequality}} &
\multicolumn{2}{c}{\textbf{Number Theory}} &
\multicolumn{1}{c}{\textbf{Combinatorics}} \\

\cmidrule(lr){2-4}\cmidrule(lr){5-6}\cmidrule(l){7-7}

& {567NEQ (92)} & {ChenNEQ (42)} & {MO-INT (20)} &
{PutnamBench (66)} & {ProverBench (40)} &
{CombiBench (43)}
\\

\midrule
\multicolumn{7}{l}{\textbf{Proprietary LLMs}} \\
\addlinespace[2pt]

\hspace{0.8em}GPT-5.3-Codex & 21 (22.8\%) & 27 (64.3\%) & 8 (40.0\%) & 3 (4.5\%) & 12 (30.0\%) & 12 (27.9\%) \\
\hspace{0.8em}Claude 4.6 Opus & 14 (15.2\%) & 19 (45.2\%) & 6 (30.0\%) & 0 (0.0\%) & 11 (27.5\%) & 12 (27.9\%) \\
\hspace{0.8em}Gemini 2.5 Pro & 3 (3.3\%) & 5 (11.9\%) & 1 (5.0\%) & 0 (0.0\%) & 5 (12.5\%) & 2 (4.7\%) \\
\hspace{0.8em}Gemini 3.1 Pro & 44 (47.8\%) & 33 (78.6\%) & 15 (75.0\%) & 4 (6.1\%) & 13 (32.5\%) & 7 (16.3\%) \\

\addlinespace[4pt]
\midrule

\multicolumn{7}{l}{\textbf{Open-source LLMs}} \\
\addlinespace[2pt]

\hspace{0.8em}DeepSeek-Prover-V2-7B & 7 (7.6\%) & 13 (31.0\%) & 3 (15.0\%) & 0 (0.0\%) & 10 (25.0\%) & 8 (18.6\%) \\
\hspace{0.8em}Goedel-Prover-V2-8B & 9 (9.8\%) & 20 (47.6\%) & 4 (20.0\%) & 0 (0.0\%) & 10 (25.0\%) & 7 (16.3\%) \\
\hspace{0.8em}Goedel-Prover-V2-32B & 25 (27.2\%) & 26 (61.9\%) & 10 (50.0\%) & 4 (6.1\%) & 12 (30.0\%) & 13 (30.2\%) \\

\addlinespace[4pt]
\midrule

\multicolumn{7}{l}{\textbf{Agentic System}} \\
\addlinespace[2pt]

\hspace{0.8em}Hilbert (GPT-5.3-Codex) & 47 (51.1\%) & 30 (71.4\%) & 12 (60.0\%) & 3 (4.5\%) & 13 (32.5\%) & 17 (39.5\%) \\
\hspace{0.8em}Hilbert (Gemini 2.5 Pro) & 36 (39.1\%) & 23 (54.8\%) & 12 (60.0\%) & 7 (10.6\%) & 11 (27.5\%) & 18 (41.9\%) \\
\hspace{0.8em}Hilbert (Gemini 3.1 Pro) & 51 (55.4\%) & 31 (73.8\%) & 13 (65.0\%) & 8 (12.1\%) & 16 (40.0\%) & 17 (39.5\%) \\

\addlinespace[4pt]
\midrule




\multicolumn{7}{l}{\textbf{Lemma Learning System}} \\
\addlinespace[2pt]

\hspace{0.8em}DreamProver (GPT-5.3-Codex) & 55 (59.8\%) & 33 (78.6\%) & 16 (80.0\%) & 16 (24.2\%) & 21 (52.5\%) & 26 (60.5\%) \\
\hspace{0.8em}DreamProver (Gemini 2.5 Pro) & 49 (53.3\%) & 28 (66.7\%) & 14 (70.0\%) & 16 (24.2\%) & \textbf{25 (62.5\%)} & 25 (58.1\%) \\
\hspace{0.8em}DreamProver (Gemini 3.1 Pro) & \textbf{57 (62.0\%)} & \textbf{36 (85.7\%)} & \textbf{17 (85.0\%)} & \textbf{19 (28.8\%)} & \textbf{25 (62.5\%)} & \textbf{27 (62.8\%)} \\

\bottomrule
\end{tabular}
}
\caption{
Solved problems and accuracy on three well-represented domain benchmarks.
}
\label{tab:benchmark_ID}
\end{table*}

\smallsec{Main results} Table~\ref{tab:benchmark_ID} reports performance on well-represented domains. Compared with all three categories of baselines, \framework{} consistently achieves substantial improvements across different backbone LLMs. Compared to the prior state-of-the-art system Hilbert, \framework{} improves performance by $20\%$, $114\%$, and $50\%$ on inequalities, number theory, and combinatorics, respectively, yielding an average gain of $61\%$ across domains. Notably, while Hilbert combines Goedel-Prover-V2 with proprietary LLMs to enhance reasoning, \framework{} relies solely on off-the-shelf LLMs yet achieves consistently strong performance, highlighting the effectiveness of the learned lemma library.

\subsection{RQ2: Inference Efficiency and Proof Quality}
\label{exp:cost}


\begin{figure*}[t]
  \centering
  \includegraphics[width=0.9\textwidth]{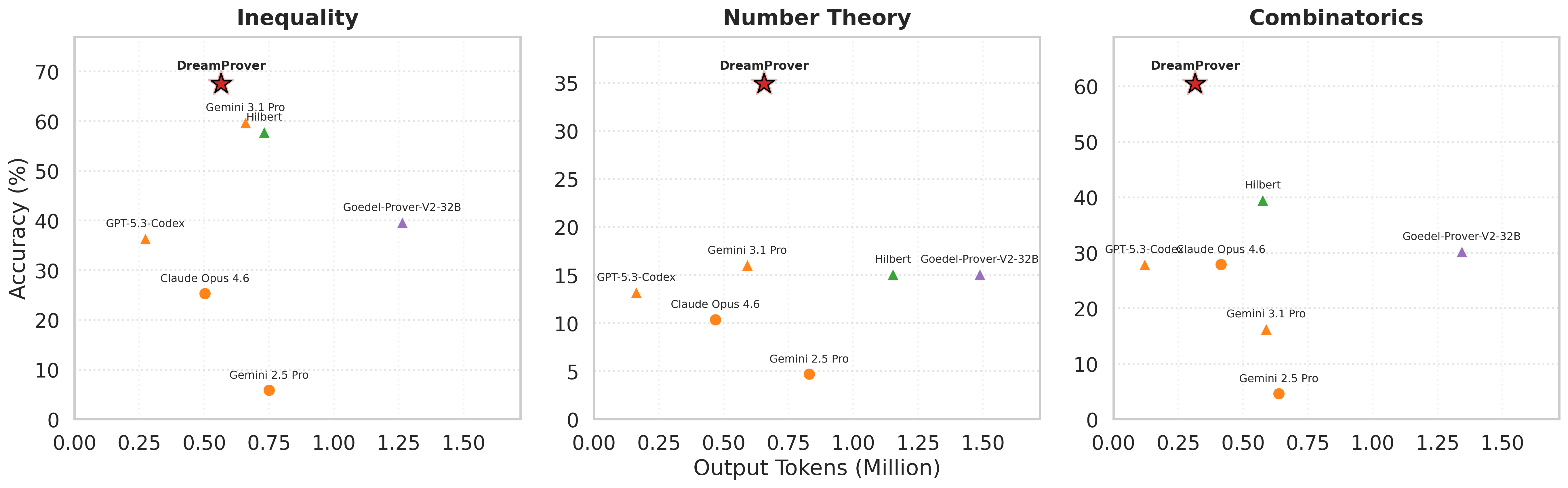}
  \caption{Total number of output tokens (millions) per sample on three well-represented domain benchmarks.}
  \label{fig:token_stats}
\end{figure*}

\smallsec{Inference efficiency}
We evaluate inference efficiency across different methods by measuring output token usage per sample, including both invisible thinking tokens and visible outputs. For both Hilbert and \framework{}, we use GPT-5.3-Codex as the base model. Figure~\ref{fig:token_stats} illustrates the relationship between token usage and the number of theorems proved across the three domains.
The results show that \framework{} achieves substantially stronger proving performance while using fewer tokens than both open-source models and agentic systems. Specifically, compared to Goedel-Prover-V2, \framework{} reduces output token usage by $55\%$, $56\%$, and $76\%$ across the three domains, respectively. Compared to Hilbert, the reductions are $42\%$, $50\%$, and $53\%$. On average, \framework{} saves $62\%$ and $48\%$ of output tokens relative to state-of-the-art open-source LLM and the agentic system, respectively.
Complete token usage statistics for all methods are reported in Table~\ref{tab:token_stats_ID} in Appendix~\ref{app:cost}.

\begin{figure*}[t]
  \centering
  \includegraphics[width=0.7\textwidth]{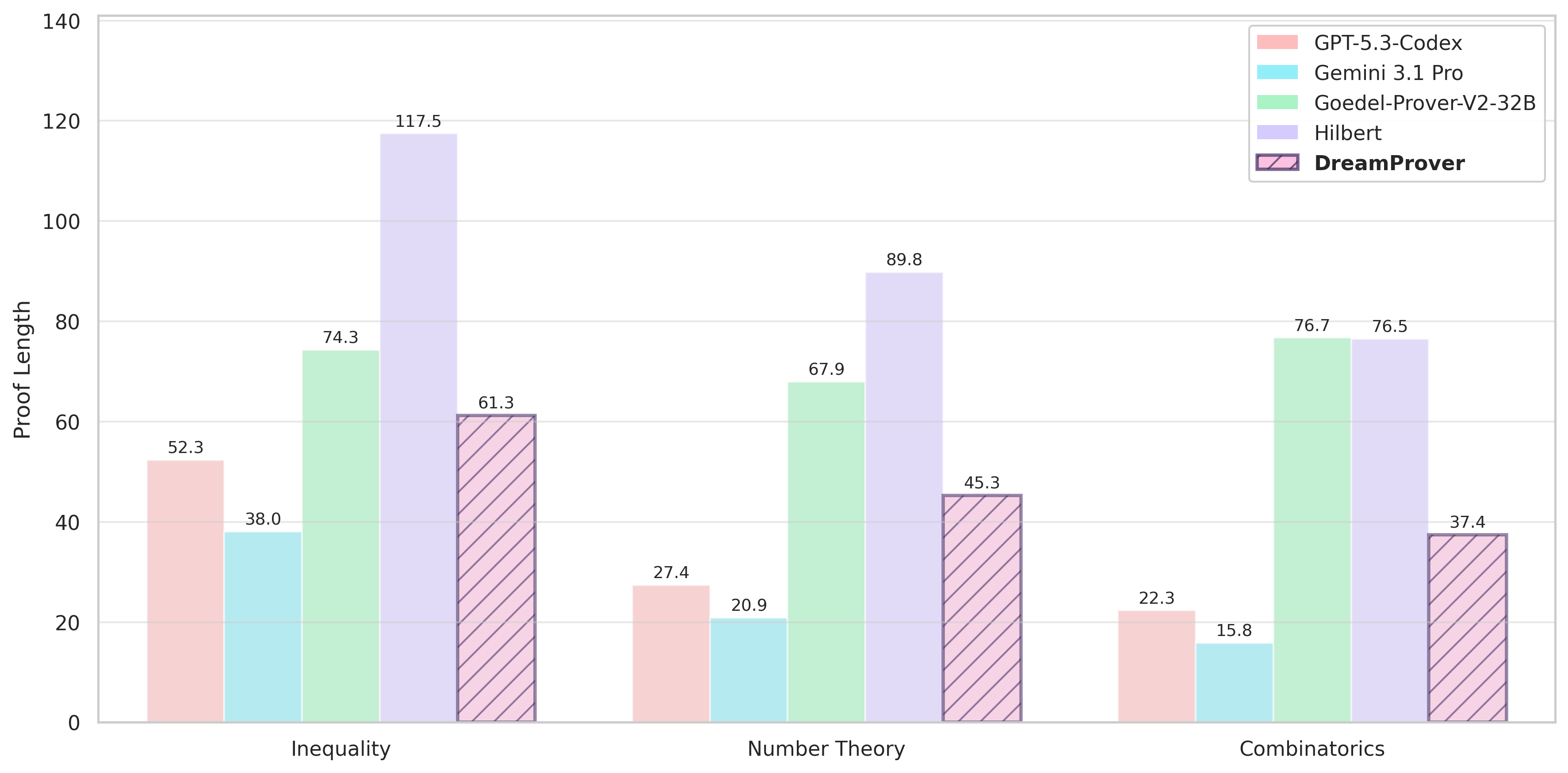}
  \caption{Proof length statistics on three well-represented domain benchmarks.}
  \label{fig:proof_length}
\end{figure*}

\smallsec{Proof Quality} We use proof length as a proxy for proof quality, as shorter and more structured proofs are generally more readable and preferred by humans. We compare \framework{} against proprietary reasoning models, Goedel-Prover-V2-32B, and the agentic system Hilbert, using GPT-5.3-Codex as the backbone model. Figure~\ref{fig:proof_length} reports the average proof lengths across the three domains. Compared to Goedel-Prover-V2, \framework{} reduces proof length by $12\%$, $33\%$, and $51\%$, respectively; compared to Hilbert, the reductions are $48\%$, $51\%$, and $51\%$. On average, this corresponds to reductions of $32\%$ and $50\%$ relative to the two baselines. Compared to proprietary LLMs, \framework{} produces longer proofs on average across all solved problems, primarily because it solves more difficult instances that require longer proofs. In Appendix~\ref{app:proof_examples}, we provide examples showing that \framework{} effectively leverages learned lemmas on unseen theorems to produce more compact proofs. Complete statistics are reported in Table~\ref{tab:proof_length_ID} in Appendix~\ref{app:cost}.

\subsection{RQ3: Effectiveness of Learned Lemmas}

\begin{table*}[ht!]
\centering
\resizebox{\textwidth}{!}{
\begin{tabular}{l cccccc}
\toprule



\textbf{Domain} & \textbf{Lemma Usage} & \textbf{Used Lemmas} & \textbf{Total Lemmas} & \textbf{Covered Theorems} & \textbf{Proved Theorems} & \textbf{Total Theorems} \\
\midrule

Inequality & 93 & 35 & 41 & 71 & 104 & 154  \\
Number Theory & 37 & 29 & 75 & 29 & 37 & 104 \\
Combinatorics & 23 & 19 & 39 & 18 & 26 & 43 \\

\bottomrule
\end{tabular}
}
\caption{Lemma library statistics on three well-represented domain benchmarks.}
\label{tab:lemma_stats}
\end{table*}

Previous work~\citep{lego_fail} shows that lemma reuse is extremely rare in prior library-learning approaches such as LEGO-Prover~\citep{lego}. In contrast, we find strong evidence that \framework{} effectively reuses lemmas from the learned library. Table~\ref{tab:lemma_stats} reports total lemma usage, the number of distinct reused lemmas, and the number of theorems covered by these lemmas in each domain. On average, $58\%$ of the lemmas generated from the training sets are reused on the test sets, contributing to proofs for $71\%$ of all successfully proved theorems. These statistics provide clear evidence that \framework{} can produce transferable lemmas that generalize to previously unseen problems.

\begin{table*}[ht!]
\centering
\resizebox{0.7\textwidth}{!}{
\begin{tabular}{l cccc}
\toprule

\textbf{Method} & \textbf{567NEQ} & \textbf{ChenNEQ} & \textbf{MO-INT} & \textbf{Total} \\
\midrule

DreamProver & 55 & 33 & 16 & 104  \\
- library optimization & 31 & 30 & 15 & 76 \\
- cluster evolving & 21 & 23 & 9 & 53 \\
\midrule
No library & 23 & 24 & 8 & 55 \\
+ retrieval (LEGO-Prover) & 27 & 25 & 9 & 61 \\

\bottomrule
\end{tabular}
}
\caption{Ablation of lemma learning strategies.}
\label{tab:learning_ablation}
\end{table*}

We further conduct ablation studies to validate the key design choices of \framework{}. Table~\ref{tab:learning_ablation} reports results on three inequality benchmarks. We first study variants that remove library optimization, i.e., iterative wake-sleep refinement, and cluster-based lemma evolution. Without iterative refinement, lemma evolution is performed in a single pass, causing the library to capture only low-level decomposed theorems. This leads to degenerate lemmas during the sleep stage and fails to optimize lemma reusability during the wake stage, allowing non-reusable lemmas to persist. As a result, the number of solved problems drops from 104 to 76.
Without semantic clustering, the system fails to capture relationships among decomposed theorems and instead treats them independently, resulting in an oversized lemma library from which useful lemmas are difficult to retrieve. The excessive library size further reduces performance from 76 to 53 solved problems, even below the vanilla inference pipeline without a library.
We further evaluate a no-library baseline and a single-pass lemma evolution method with semantic retrieval, similar to LEGO-Prover~\citep{lego}. These results show that maintaining a compact library of reusable, abstract lemmas is more effective than retrieving from a large, low-level library.

\subsection{RQ4: Underrepresented Domain Generalization}

We also extend our evaluation beyond domains covered by existing formal libraries and LLM knowledge. Specifically, we consider two additional domains: plane geometry and machine learning theory. Both domains involve problems defined by specialized axioms and theorems that are not included in mathlib.

\smallsec{Datasets} For plane geometry, we use LeanGeo-Bench~\citep{leangeo}, a competition-level benchmark formalized in Lean 4, and report results on its five subsets: UniGeo~\citep{unigeo}, library, synthetic problems, high school competitions, and Olympiad.
For machine learning theory, we evaluate on FormalML~\citep{formalml}, which covers foundational topics such as optimization and statistical learning theory. FormalML is constructed from the Optlib~\citep{li2024formalization} and FoML~\citep{sonoda2025lean} libraries, and its problems are grouped into three subsets based on ground-truth proof length.
We learn the lemma library for plane geometry using the IMO subset of LeanGeo-Bench. For machine learning theory, the library is evolved on the level-2 subset of FormalML and evaluated on the level-3 subset.

\smallsec{Baselines} Since open-source prover models are not trained on datasets that cover these domains and do not achieve comparable performance, even when provided with additional context, and since Hilbert also relies heavily on such models (e.g., Goedel-Prover-V2), we primarily compare \framework{} against proprietary reasoning LLMs. For these baselines, we provide additional contextual information, including relevant axioms and theorems, as part of the input.
Our baselines include GPT-5.3-Codex, Gemini 2.5 Pro, and Gemini 3.1 Pro Preview, all evaluated under the pass@32 setting. For \framework{}, we use the same parameter settings as in the previous evaluation.

\begin{table*}[ht!]
\centering
\resizebox{\textwidth}{!}{
\begin{tabular}{l ccccc c}
\toprule
\multirow{2}{*}{\textbf{Method}} &
\multicolumn{5}{c}{\textbf{Plane Geometry}} &
\multicolumn{1}{c}{\textbf{Machine Learning Theory}} \\

\cmidrule(lr){2-6}\cmidrule(lr){7-7}

& {UniGeo (10)} & {Library (10)} & {Synthetic (20)} & {HS Competition (10)} & {Olympiad (19)} & {FormalML-Hard (122)}
\\

\midrule
\multicolumn{7}{l}{\textbf{Proprietary LLMs}} \\
\addlinespace[2pt]

\hspace{0.8em}GPT-5.3-Codex & 10 (100.0\%) & 10 (100.0\%) & 15 (75.0\%) & 1 (5.0\%) & 0 (0.0\%) & 59 (48.4\%) \\
\hspace{0.8em}Gemini 2.5 Pro & 10 (100.0\%) & 5 (50.0\%) & 12 (60.0\%) & 0 (0.0\%) & 0 (0.0\%) & 25 (20.5\%) \\
\hspace{0.8em}Gemini 3.1 Pro & 10 (100.0\%) & 7 (70.0\%) & 13 (65.0\%) & 2 (10.0\%) & 1 (5.3\%) & 33 (27.0\%) \\

\addlinespace[4pt]
\midrule




\multicolumn{7}{l}{\textbf{Lemma Learning System}} \\
\addlinespace[2pt]

\hspace{0.8em}DreamProver (GPT-5.3-Codex) & \textbf{10 (100.0\%)} & \textbf{10 (100.0\%)} & \textbf{17 (85.0\%)} & 7 (70.0\%) & 7 (36.8\%) & \textbf{95 (77.9\%)} \\
\hspace{0.8em}DreamProver (Gemini 2.5 Pro) & \textbf{10 (100.0\%)} & \textbf{10 (100.0\%)} & 16 (80.0\%) & 7 (70.0\%) & \textbf{8 (42.1\%)} & 85 (69.7\%) \\
\hspace{0.8em}DreamProver (Gemini 3.1 Pro) & \textbf{10 (100.0\%)} & \textbf{10 (100.0\%)} & \textbf{17 (85.0\%)} & \textbf{9 (90.0\%)} & 7 (36.8\%) & 93 (76.2\%) \\

\bottomrule
\end{tabular}
}
\caption{Solved problems and accuracy on two under-represented domain benchmarks.}
\label{tab:benchmark_OOD}
\end{table*}

\smallsec{Main Results} Table~\ref{tab:benchmark_OOD} shows that \framework{} continues to benefit from the learned lemma library in new domains, consistently achieving state-of-the-art performance across all datasets and backbone LLMs. Compared to proprietary LLM baselines, it achieves relative improvements of $64\%$ and $161\%$ on the two domains.
Notably, on challenging plane geometry problems from high school competition and Olympiad benchmarks, where LLMs alone solve only 0-3 instances, \framework{} significantly improves performance, solving up to 14-16 instances with the learned lemma library.
We further report detailed token and proof-length statistics in Tables~\ref{tab:token_stats_OOD} and~\ref{tab:proof_length_OOD} in Appendix~\ref{app:cost}. These results demonstrate that \framework{} generalizes beyond domains covered by LLM knowledge and can be applied to a broader range of domains and applications beyond the mathlib library.
\section{Limitations and Future Work}
\label{limitation}

While the learned lemma libraries in each domain are currently small and easily fit within the context window of proprietary LLMs, adapting to smaller models with limited context may be challenging - especially when the problem domain is unknown and multiple libraries may need to be included, leading to longer inputs. A practical direction is to develop stronger premise selection models~\citep{leansearch,tbps,lu2025lean} that retrieve relevant lemmas before passing them to the LLM. Moreover, in settings without sufficient training data for a specific domain (especially in mathematical research), \framework{} can operate in an online learning fashion, incrementally accumulating useful lemmas during evaluation to support future proofs. We leave these directions for future work.
\section{Conclusion}
\label{conclusion}
We present \framework{}, a theorem-proving agent that learns reusable lemma libraries via iterative wake-sleep cycles: it proves problems and discovers lemmas in the wake phase, and abstracts and refines them in the sleep phase. Across diverse mathematical domains, \framework{} effectively reuses the learned libraries, significantly increasing the number of theorems proved while producing more concise, structured proofs with lower token cost.

\clearpage

\bibliography{reference}
\bibliographystyle{colm2026_conference}

\newpage
\appendix
\onecolumn

\section{Implementation Details}
\label{app:implementation_details}

On each domain, we train \framework{} for 5 wake-sleep cycles, with maximum decomposition depth = 3, sketch generation attempts = 1, direct proof generation attempts = 4, error correction for sketch generation and direct proof generation = 6. During inference, we set the maximum decomposition depth = 1 to allow only sketch-and-prove. We set sketch generation attempts = 4, direct proof generation attempts = 4, error correction for sketch generation and direct proof generation = 6. For the Hilbert baseline, we set the default parameters except for maximum recursion depth = 2 since the compute budget grows exponentially with larger recursion depth.

\section{Output token statistics}
\label{app:cost}

In Table~\ref{tab:token_stats_ID} and~\ref{tab:token_stats_OOD}, we list all the output token usage of the evaluated methods across the five domain. In Table~\ref{tab:training_cost}, we further compare the training output token usage accumulated on 5 iteration per sample and the inference output token usage per sample. The training token usage is shown to be roughly equal to or less than the inference token usage. Since the training set contains only equal or less than 100 samples, and the library is learned only once on each domain, the total cost of library learning is controlled within a reasonable range.

\begin{table*}[ht!]
\centering
\resizebox{\textwidth}{!}{
\begin{tabular}{l ccc cc c}
\toprule
\multirow{2}{*}{\textbf{Method}} &
\multicolumn{3}{c}{\textbf{Inequality}} &
\multicolumn{2}{c}{\textbf{Number Theory}} &
\multicolumn{1}{c}{\textbf{Combinatorics}} \\

\cmidrule(lr){2-4}\cmidrule(lr){5-6}\cmidrule(l){7-7}

& {567NEQ (92)} & {ChenNEQ (42)} & {MO-INT (20)} &
{PutnamBench (66)} & {ProverBench (40)} &
{CombiBench (43)}
\\

\midrule
\multicolumn{7}{l}{\textbf{Proprietary LLMs}} \\
\addlinespace[2pt]

\hspace{0.8em}GPT-5.3-Codex & 0.28 & 0.24 & 0.30 & 0.18 & 0.13 & 0.12 \\
\hspace{0.8em}Claude 4.6 Opus & 0.51 & 0.47 & 0.52 & 0.51 & 0.40 & 0.41 \\
\hspace{0.8em}Gemini 2.5 Pro & 0.81 & 0.65 & 0.68 & 0.96 & 0.61 & 0.64 \\
\hspace{0.8em}Gemini 3.1 Pro & 0.62 & 0.68 & 0.76 & 0.59 & 0.60 & 0.59 \\

\addlinespace[4pt]
\midrule

\multicolumn{7}{l}{\textbf{Open-source LLMs}} \\
\addlinespace[2pt]

\hspace{0.8em}DeepSeek-Prover-V2-7B & 0.38 & 0.33 & 0.36 & 0.58 & 0.50 & 0.42 \\
\hspace{0.8em}Goedel-Prover-V2-8B & 1.44 & 0.61 & 0.82 & 2.10 & 1.72 & 1.96 \\
\hspace{0.8em}Goedel-Prover-V2-32B & 1.52 & 0.79 & 1.07 & 1.62 & 1.27 & 1.31 \\

\addlinespace[4pt]
\midrule

\multicolumn{7}{l}{\textbf{Agentic System}} \\
\addlinespace[2pt]

\hspace{0.8em}Hilbert (GPT-5.3-Codex) & 0.78 & 0.52 & 0.96 & 1.20 & 1.07 & 0.58 \\
\hspace{0.8em}Hilbert (Gemini 2.5 Pro) & 2.86 & 1.48 & 3.40 & 2.51 & 2.19 & 1.89 \\
\hspace{0.8em}Hilbert (Gemini 3.1 Pro) & 3.10 & 1.28 & 2.92 & 1.87 & 1.52 & 2.05 \\

\addlinespace[4pt]
\midrule




\multicolumn{7}{l}{\textbf{Lemma Learning System}} \\
\addlinespace[2pt]

\hspace{0.8em}DreamProver (GPT-5.3-Codex) & 0.53 & 0.68 & 0.50 & 0.89 & 0.27 & 0.32 \\
\hspace{0.8em}DreamProver (Gemini 2.5 Pro) & 1.12 & 1.39 & 1.20 & 1.13 & 0.54 & 0.75 \\
\hspace{0.8em}DreamProver (Gemini 3.1 Pro) & 1.30 & 1.47 & 0.92 & 1.15 & 0.64 & 0.93 \\

\bottomrule
\end{tabular}
}
\caption{Token statistics (thousands of output tokens) on three well-represented domain benchmarks.}
\label{tab:token_stats_ID}
\end{table*}

\begin{table*}[ht!]
\centering
\resizebox{0.9\textwidth}{!}{
\begin{tabular}{l ccccc c}
\toprule
\multirow{2}{*}{\textbf{Method}} &
\multicolumn{5}{c}{\textbf{Plane Geometry}} &
\multicolumn{1}{c}{\textbf{Machine Learning Theory}} \\

\cmidrule(lr){2-6}\cmidrule(lr){7-7}

& {UniGeo (10)} & {Library (10)} & {Synthetic (20)} & {HS Competition (10)} & {Olympiad (19)} & {FormalML-Hard (122)}
\\

\midrule
\multicolumn{7}{l}{\textbf{Proprietary LLMs}} \\
\addlinespace[2pt]

\hspace{0.8em}GPT-5.3-Codex & 0.12 & 0.08 & 0.14 & 0.19 & 0.28 & 0.10 \\
\hspace{0.8em}Gemini 3.1 Pro & 0.21 & 0.25 & 0.38 & 0.39 & 0.56 & 0.28 \\
\hspace{0.8em}Gemini 2.5 Pro & 0.18 & 0.28 & 0.34 & 0.32 & 0.66 & 0.29 \\

\addlinespace[4pt]
\midrule




\multicolumn{7}{l}{\textbf{Lemma Learning System}} \\
\addlinespace[2pt]

\hspace{0.8em}DreamProver (GPT-5.3-Codex) & 0.10 & 0.14 & 0.12 & 0.16 & 0.23 & 0.23 \\
\hspace{0.8em}DreamProver (Gemini 2.5 Pro) & 0.32 & 0.20 & 0.22 & 0.22 & 0.40 & 0.36 \\
\hspace{0.8em}DreamProver (Gemini 3.1 Pro) & 0.25 & 0.27 & 0.27 & 0.18 & 0.39 & 0.40 \\

\bottomrule
\end{tabular}
}
\caption{Token statistics (thousands of output tokens) on two under-represented domain benchmarks.}
\label{tab:token_stats_OOD}
\end{table*}

\begin{table*}[ht!]
\centering
\resizebox{0.9\textwidth}{!}{
\begin{tabular}{l ccccc}
\toprule

\textbf{} & \textbf{Inequality} & \textbf{Number Theory} & \textbf{Combinatorics} & \textbf{Plane Geometry} & \textbf{Machine Learning Theory} \\
\midrule

Hilbert & 0.73 & 1.15 & 0.58 & - & - \\
DreamProver (Training) & 0.29 & 0.44 & 0.22 & 0.18 & 0.16 \\
DreamProver (Inference) & 0.51 & 0.67 & 0.32 & 0.15 & 0.28 \\

\bottomrule
\end{tabular}
}
\caption{Training token statistics (thousands of output tokens) on all evaluated domains.}
\label{tab:training_cost}
\end{table*}

\pagebreak
\section{Proof Length}

In Table~\ref{tab:proof_length_ID} and~\ref{tab:proof_length_OOD}, we list all the proof lengths of the evaluated methods across the five domains.

\begin{table*}[ht!]
\centering
\resizebox{\textwidth}{!}{
\begin{tabular}{l ccc cc c}
\toprule
\multirow{2}{*}{\textbf{Method}} &
\multicolumn{3}{c}{\textbf{Inequality}} &
\multicolumn{2}{c}{\textbf{Number Theory}} &
\multicolumn{1}{c}{\textbf{Combinatorics}} \\

\cmidrule(lr){2-4}\cmidrule(lr){5-6}\cmidrule(l){7-7}

& {567NEQ (92)} & {ChenNEQ (42)} & {MO-INT (20)} &
{PutnamBench (66)} & {ProverBench (40)} &
{CombiBench (43)}
\\

\midrule
\multicolumn{7}{l}{\textbf{Proprietary LLMs}} \\
\addlinespace[2pt]

\hspace{0.8em}GPT-5.3-Codex & 56.6 & 46.6 & 60.5 & 85.7 & 12.8 & 22.3 \\
\hspace{0.8em}Claude 4.6 Opus & 15.8 & 15.2 & 17.0 & - & 5.5 & 8.0 \\
\hspace{0.8em}Gemini 2.5 Pro & 22.3 & 25.9 & 35.0 & - & 12.7 & 23.8 \\
\hspace{0.8em}Gemini 3.1 Pro & 40.9 & 29.7 & 47.9 & 67.3 & 6.6 & 15.8 \\

\addlinespace[4pt]
\midrule

\multicolumn{7}{l}{\textbf{Open-source LLMs}} \\
\addlinespace[2pt]

\hspace{0.8em}DeepSeek-Prover-V2-7B & 10.1 & 14.5 & 61.8 & - & 19.6 & 41.4 \\
\hspace{0.8em}Goedel-Prover-V2-8B & 24.1 & 21.8 & 76.0 & - & 21.7 & 45.3 \\
\hspace{0.8em}Goedel-Prover-V2-32B & 64.4 & 63.3 & 128.0 & 146.2 & 41.8 & 76.7 \\

\addlinespace[4pt]
\midrule

\multicolumn{7}{l}{\textbf{Agentic System}} \\
\addlinespace[2pt]

\hspace{0.8em}Hilbert (GPT-5.3-Codex) & 118.6 & 113.4 & 124.1 & 157.3 & 67.3 & 76.5 \\
\hspace{0.8em}Hilbert (Gemini 2.5 Pro) & 97.1 & 79.5 & 93.4 & 126.9 & 45.8 & 85.7 \\
\hspace{0.8em}Hilbert (Gemini 3.1 Pro) & 85.8 & 87.9 & 132.6 & 129.4 & 65.7 & 63.4 \\

\addlinespace[4pt]
\midrule




\multicolumn{7}{l}{\textbf{Lemma Learning System}} \\
\addlinespace[2pt]

\hspace{0.8em}DreamProver (GPT-5.3-Codex) & 65.2 & 57.6 & 55.3 & 78.9 & 34.1 & 37.4 \\
\hspace{0.8em}DreamProver (Gemini 2.5 Pro) & 43.8 & 32.9 & 45.3 & 62.5 & 41.8 & 29.6 \\
\hspace{0.8em}DreamProver (Gemini 3.1 Pro) & 60.3 & 37.8 & 50.6 & 83.2 & 51.7 & 31.3 \\

\bottomrule
\end{tabular}
}
\caption{Proof length statistics on three well-represented domain benchmarks.}
\label{tab:proof_length_ID}
\end{table*}

\begin{table*}[ht!]
\centering
\resizebox{0.9\textwidth}{!}{
\begin{tabular}{l ccccc c}
\toprule
\multirow{2}{*}{\textbf{Method}} &
\multicolumn{5}{c}{\textbf{Plane Geometry}} &
\multicolumn{1}{c}{\textbf{Machine Learning Theory}} \\

\cmidrule(lr){2-6}\cmidrule(lr){7-7}

& {UniGeo (10)} & {Library (10)} & {Synthetic (20)} & {HS Competition (10)} & {Olympiad (19)} & {FormalML-Hard (122)}
\\

\midrule
\multicolumn{7}{l}{\textbf{Proprietary LLMs}} \\
\addlinespace[2pt]

\hspace{0.8em}GPT-5.3-Codex & 4.1 & 12.5 & 14.5 & 32.0 & - & 11.2 \\
\hspace{0.8em}Gemini 2.5 Pro & 5.6 & 10.0 & 16.9 & - & - & 32.8 \\
\hspace{0.8em}Gemini 3.1 Pro & 5.2 & 17.3 & 19.2 & 41.5 & 62.0 & 39.7 \\

\addlinespace[4pt]
\midrule




\multicolumn{7}{l}{\textbf{Lemma Learning System}} \\
\addlinespace[2pt]

\hspace{0.8em}DreamProver (GPT-5.3-Codex) & 7.0 & 10.4 & 17.3 & 22.3 & 28.0  & 23.3 \\
\hspace{0.8em}DreamProver (Gemini 2.5 Pro) & 7.3 & 13.8 & 15.4 & 23.3 & 36.3 & 27.5\\
\hspace{0.8em}DreamProver (Gemini 3.1 Pro) & 6.2 & 17.3 & 21.1 & 17.8 & 31.4 & 31.6 \\

\bottomrule
\end{tabular}
}
\caption{Proof length statistics on two under-represented domain benchmarks.}
\label{tab:proof_length_OOD}
\end{table*}

\section{Lemma Library}
\label{app:lemma_library}

We list 10 representative lemmas from the library in each domain to show the diversity and quality of the evolved lemmas.

\subsection{Inequality}

\begin{minted}[
    fontsize=\scriptsize,
    breaklines,
    breakanywhere,
    frame=lines,
    framesep=2mm,
    tabsize=2,
    samepage=false  % Allow page breaks
]{lean}
import Mathlib
import Aesop

set_option maxHeartbeats 0

open BigOperators Real Nat Topology Rat

theorem am_gm_inequality_three_vars
    (x y z : ℝ) (hx : 0 < x) (hy : 0 < y) (hz : 0 < z) :
    (x * y * z) ^ ((1 : ℝ) / 3) ≤ (x + y + z) / (3 : ℝ) := by <proof>

theorem am_hm_inequality_three_vars
    (x y z : ℝ) (hx : 0 < x) (hy : 0 < y) (hz : 0 < z) :
    (9 : ℝ) ≤ (x + y + z) * (1 / x + 1 / y + 1 / z) := by <proof>

theorem cauchy_schwarz_inequality_three_vars
    (a₁ a₂ a₃ b₁ b₂ b₃ : ℝ) :
    (a₁ * b₁ + a₂ * b₂ + a₃ * b₃) ^ 2 ≤ (a₁ ^ 2 + a₂ ^ 2 + a₃ ^ 2) * (b₁ ^ 2 + b₂ ^ 2 + b₃ ^ 2) := by <proof>

theorem titu_lemma_three_reals
    {a b c x y z : ℝ}
    (ha : 0 < a) (hb : 0 < b) (hc : 0 < c) :
    (x + y + z) ^ 2 / (a + b + c) ≤ x ^ 2 / a + y ^ 2 / b + z ^ 2 / c := by <proof>

theorem schur_inequality_cubic_reals
    {a b c : ℝ}
    (ha : 0 ≤ a) (hb : 0 ≤ b) (hc : 0 ≤ c) :
    a ^ 3 + b ^ 3 + c ^ 3 + 3 * a * b * c ≥ a * b * (a + b) + b * c * (b + c) + c * a * (c + a) := by <proof>

theorem holder_inequality_two_sequences
    {a b c x y z p q : ℝ}
    (ha : 0 ≤ a) (hb : 0 ≤ b) (hc : 0 ≤ c) (hx : 0 ≤ x) (hy : 0 ≤ y) (hz : 0 ≤ z)
    (hp : 1 < p) (hpq : (1 : ℝ) / p + (1 : ℝ) / q = 1) :
    a * x + b * y + c * z ≤ (a ^ p + b ^ p + c ^ p) ^ ((1 : ℝ) / p) * (x ^ q + y ^ q + z ^ q) ^ ((1 : ℝ) / q) := by <proof>

theorem nesbitt_inequality
    {a b c : ℝ} (ha : 0 < a) (hb : 0 < b) (hc : 0 < c) :
    a / (b + c) + b / (c + a) + c / (a + b) ≥ (3 : ℝ) / 2 := by <proof>

theorem weighted_am_gm_three_reals
    {x y z w₁ w₂ w₃ : ℝ}
    (hx : 0 ≤ x) (hy : 0 ≤ y) (hz : 0 ≤ z)
    (hw₁ : 0 < w₁) (hw₂ : 0 < w₂) (hw₃ : 0 < w₃) (hw_sum : w₁ + w₂ + w₃ = 1) :
    x ^ w₁ * y ^ w₂ * z ^ w₃ ≤ w₁ * x + w₂ * y + w₃ * z := by <proof>

theorem newton_inequality_S2_S1S3
    {a b c : ℝ}
    (ha : 0 ≤ a) (hb : 0 ≤ b) (hc : 0 ≤ c) :
    (3 : ℝ) * (a + b + c) * (a * b * c) ≤ (a * b + b * c + c * a) ^ 2 := by <proof>

theorem rearrangement_inequality_four
    (a b c d x y z w : ℝ)
    (h_ord1 : a ≤ b ∧ b ≤ c ∧ c ≤ d)
    (h_ord2 : x ≤ y ∧ y ≤ z ∧ z ≤ w) :
    a * w + b * z + c * y + d * x ≤ a * x + b * y + c * z + d * w := by <proof>
\end{minted}

\subsection{Number Theory}

\begin{minted}[
    fontsize=\scriptsize,
    breaklines,
    breakanywhere,
    frame=lines,
    framesep=2mm,
    tabsize=2,
    samepage=false  % Allow page breaks
]{lean}
import Mathlib
import Aesop

set_option maxHeartbeats 0

open BigOperators Real Nat Topology Rat

theorem binomial_ratio_identity : ∀ n k : ℕ, k > 0 → k ≤ n → (k : ℤ) * (Nat.choose n k : ℤ) = ((n : ℤ) - (k : ℤ) + (1 : ℤ)) * (Nat.choose n (k - 1) : ℤ) := by <proof>

theorem euclids_lemma_for_primes {p a b : ℤ} : Prime p → p ∣ a * b → p ∣ a ∨ p ∣ b := by <proof>

theorem gcd_of_shifted_integer {n k : ℤ} : Int.gcd n (n + k) ∣ k.natAbs := by <proof>

theorem infinitude_of_primes :
  Set.Infinite {p : ℕ | Nat.Prime p} := by <proof>

theorem dirichlet_theorem_on_arithmetic_progressions (a n : ℕ) (h : Nat.Coprime a n) :
  Set.Infinite {p : ℕ | Nat.Prime p ∧ p ≡ a [MOD n]} := by <proof>

theorem coprime_iff_forall_prime_not_dvd (a n : ℤ) (hn' : n.natAbs ≠ 0) :
  Nat.gcd a.natAbs n.natAbs = 1 ↔ ∀ p : ℕ, p.Prime → (p : ℤ) ∣ n → ¬ ((p : ℤ) ∣ a) := by <proof>

theorem fermat_little_theorem {p a : ℤ} : p > 0 → Prime p → ¬ (p ∣ a) → a ^ (p - (1 : ℤ)).toNat % p = (1 : ℤ) % p := by <proof>

theorem euler_criterion {p a : ℤ} : p > 2 → Prime p → ¬ (p ∣ a) → ((∃ x : ℤ, x^2 % p = a % p) ↔ a ^ ((p - (1 : ℤ)) / (2 : ℤ)).toNat % p = (1 : ℤ) % p) := by <proof>

theorem coprime_factor_square_prime_decomposition {a b p k : ℤ} : a > 0 → b > 0 → Prime p → Int.gcd a b = 1 → a * b = p * k^2 → ((∃ r s : ℤ, a = p * r^2 ∧ b = s^2) ∨ (∃ r s : ℤ, a = r^2 ∧ b = p * s^2)) := by <proof>

theorem card_of_units_zmod (p : ℕ) [hp : Fact p.Prime] : Fintype.card (ZMod p)ˣ = p - 1 := by <proof>

theorem val_of_pow_in_units_zmod (p : ℕ) [Fact p.Prime] (u : (ZMod p)ˣ) (k : ℕ) : (u^k : ZMod p) = (u : ZMod p)^k := by <proof>
\end{minted}

\subsection{Combinatorics}

\begin{minted}[
    fontsize=\scriptsize,
    breaklines,
    breakanywhere,
    frame=lines,
    framesep=2mm,
    tabsize=2,
    samepage=false  % Allow page breaks
]{lean}
import Mathlib
import Aesop

open BigOperators Classical ENNReal Equiv EuclideanGeometry Filter Finset Fintype Function Lex List MeasureTheory Nat ProbabilityTheory Real SimpleGraph Real Nat Topology Rat

set_option maxHeartbeats 0

theorem choose_symm_comb {n k : ℕ} (h : k ≤ n) : Nat.choose n k = Nat.choose n (n - k) := by <proof>

theorem choose_succ_succ_comb (n k : ℕ) : Nat.choose (n + 1) (k + 1) = Nat.choose n k + Nat.choose n (k + 1) := by <proof>

theorem fib_cassini_identity : ∀ n : ℕ, n % 2 = 0 → Nat.fib (n + 1) * Nat.fib (n + 1) = Nat.fib n * Nat.fib (n + 2) + 1 := by <proof>

theorem Finset.card_eq_of_mem_powersetCard {α : Type*} {X : Finset α} {r : ℕ} {s : Finset α} :
    s ∈ X.powersetCard r → s.card = r := by <proof>

theorem Finset.card_compl_of_fintype {α : Type*} [Fintype α] (s : Finset α) :
    sᶜ.card = Fintype.card α - s.card := by <proof>

theorem card_perm_fixing_finset {α : Type*} [Fintype α] [DecidableEq α] (s : Finset α) :
    (Finset.univ.filter (fun (σ : Equiv.Perm α) => ∀ x ∈ s, σ x = x)).card = (Fintype.card α - s.card)! := by <proof>

theorem Finset.compl_congr {α : Type*} [Fintype α] {s t : Finset α} :
    s = t → sᶜ = tᶜ := by <proof>

theorem exists_unreachable_from_reachable_pair {V : Type*} (G : SimpleGraph V) (hnp : ¬ G.Preconnected) {u v : V} (huv : G.Reachable u v) :
    ∃ w : V, ¬ G.Reachable u w := by <proof>

theorem non_adjacent_of_not_reachable {V : Type*} (G : SimpleGraph V) {u v : V} (h : ¬ G.Reachable u v) :
    ¬ G.Adj u v := by <proof>

theorem reachable_in_compl_of_common_non_neighbor {V : Type*} (G : SimpleGraph V) {u v w : V} (huw : ¬ G.Adj u w) (hvw : ¬ G.Adj v w) :
    Gᶜ.Reachable u v := by <proof>
\end{minted}

\subsection{Geometry}

\begin{minted}[
    fontsize=\scriptsize,
    breaklines,
    breakanywhere,
    frame=lines,
    framesep=2mm,
    tabsize=2,
    samepage=false  % Allow page breaks
]{lean}
import Mathlib
import SystemE
import LeanGeo
namespace LeanGeo

theorem power_of_point_interior_formulation (A O I P : Point) (Ω : Circle)
  (h_is_centre : O.isCentre Ω)
  (h_A_on : A.onCircle Ω)
  (h_P_on : P.onCircle Ω)
  (h_between : between A I P) :
  |(O─A)| * |(O─A)| - |(O─I)| * |(O─I)| = |(A─I)| * |(I─P)| := by <proof>

theorem power_of_incenter_equals_2Rr (A B C O I D P : Point) (BC : Line) (Ω : Circle)
  (h_circumcenter : Circumcentre O A B C)
  (h_incenter : Incentre I A B C)
  (h_foot : Foot I D BC)
  (h_is_centre : O.isCentre Ω)
  (h_circumcircle : A.onCircle Ω ∧ B.onCircle Ω ∧ C.onCircle Ω)
  (h_P_on : P.onCircle Ω)
  (h_coll_AIP : Coll A I P)
  (h_ne : P ≠ A) :
  |(A─I)| * |(I─P)| = 2 * |(O─A)| * |(I─D)| := by <proof>

theorem line_circle_intersections_are_unique (A P Q : Point) (L : Line) (Ω : Circle)
  (h_A_on_circle : A.onCircle Ω)
  (h_A_on_line : A.onLine L)
  (h_intersections : P.onCircle Ω ∧ P.onLine L ∧ Q.onCircle Ω ∧ Q.onLine L ∧ P ≠ Q) :
  (P = A ∧ Q ≠ A) ∨ (Q = A ∧ P ≠ A) := by <proof>

theorem incenter_and_arc_midpoint_are_collinear (A B C I X : Point) (AB BC CA : Line) (Ω : Circle) :
    formAcuteTriangle A B C AB BC CA ∧
    Circumcircle Ω A B C ∧
    X.onCircle Ω ∧ |(X─B)| = |(X─C)| ∧ X.opposingSides A BC ∧
    Incentre I A B C →
    Coll A I X := by <proof>

theorem triangle_angle_sum (A B C : Point) :
    Triangle A B C →
    ∠C:A:B + ∠A:B:C + ∠B:C:A = ∟ + ∟ := by <proof>

theorem inscribed_angle_subtending_half_arc (A B C Y Z : Point) (AB BC CA : Line) (Ω : Circle) :
    formAcuteTriangle A B C AB BC CA ∧
    Circumcircle Ω A B C ∧
    Y.onCircle Ω ∧
    Z.onCircle Ω ∧ |(Z─A)| = |(Z─B)| ∧ Z.opposingSides C AB →
    ∠A:Y:Z = ∠B:C:A / 2 := by <proof>

theorem line_to_arc_midpoint_bisects_angle (A B C X : Point) (AB BC CA : Line) (Ω : Circle) :
    formAcuteTriangle A B C AB BC CA ∧
    Circumcircle Ω A B C ∧
    X.onCircle Ω ∧ |(X─B)| = |(X─C)| ∧ X.opposingSides A BC →
    ∠B:A:X = ∠C:A:X := by <proof>

theorem incenter_is_on_angle_bisector (A B C I : Point) :
    Triangle A B C →
    Incentre I A B C →
    ∠B:A:I = ∠C:A:I := by <proof>

theorem power_of_points_on_line_parallel_to_side :
  ∀ (A B C M O T : Point) (AB BC CA AT : Line) (Γ : Circle),
    formTriangle A B C AB BC CA ∧
    MidPoint B M C ∧
    distinctPointsOnLine A T AT ∧
    T.onLine AT ∧
    ¬(AT.intersectsLine BC) ∧
    Circumcircle Γ A B C ∧
    O.isCentre Γ →
    |(T─O)| * |(T─O)| - |(T─M)| * |(T─M)| = |(A─O)| * |(A─O)| - |(A─M)| * |(A─M)| := by <proof>

theorem power_property_of_orthic_related_line :
  ∀ (A B C M E F K L O T : Point) (AB BC CA KL : Line) (Γ : Circle),
    formAcuteTriangle A B C AB BC CA ∧
    MidPoint B M C ∧
    Foot B E CA ∧
    Foot C F AB ∧
    MidPoint M K E ∧
    MidPoint M L F ∧
    distinctPointsOnLine K L KL ∧
    T.onLine KL ∧
    Circumcircle Γ A B C ∧
    O.isCentre Γ →
    |(T─A)| * |(T─A)| - |(T─O)| * |(T─O)| = |(A─M)| * |(A─M)| - |(A─O)| * |(A─O)| := by <proof>
\end{minted}

\subsection{Machine Learning Theory}

\begin{minted}[
    fontsize=\scriptsize,
    breaklines,
    breakanywhere,
    frame=lines,
    framesep=2mm,
    tabsize=2,
    samepage=false  % Allow page breaks
]{lean}
import Mathlib
import FoML
import Optlib

open scoped ENNReal MeasureTheory NNReal Pointwise ProbabilityTheory Topology
open Asymptotics BCD BigOperators Bornology ContinuousLinearMap ENNReal Filter Finset Function InnerProduct InnerProductSpace LinearMap Matrix MeasureTheory Metric NNReal PNat ProbabilityTheory Real Set Topology

theorem quadratic_form_difference_identity {m n : ℕ+} (A : Matrix (Fin m) (Fin n) ℝ) (x y : EuclideanSpace ℝ (Fin n)) :
  A *ᵥ y ⬝ᵥ A *ᵥ y - A *ᵥ x ⬝ᵥ A *ᵥ x - 2 * ((Aᵀ * A) *ᵥ x ⬝ᵥ (y - x)) = ‖A *ᵥ y - A *ᵥ x‖ ^ 2 := by <proof>

theorem nesterov_core_recursion.{u_1} {E : Type u_1} [inst : NormedAddCommGroup E] [inst_1 : InnerProductSpace ℝ E] [inst_2 : CompleteSpace E] {f : E → ℝ} {f' : E → E} {x_star x₀ : E} {γ : ℕ+ → ℝ} {alg : Nesterov f f' γ x₀} (h_convex : ConvexOn ℝ univ f) (is_min : IsMinOn f univ x_star) (k : ℕ+) (hk : 1 < k) :
  1 / (γ k) ^ 2 * (f (alg.x ↑k) - f x_star) - (1 - γ k) / (γ k) ^ 2 * (f (alg.x (↑k - 1)) - f x_star) ≤
    ↑alg.l / 2 * ‖alg.v (↑k - 1) - x_star‖ ^ 2 - ↑alg.l / 2 * ‖alg.v ↑k - x_star‖ ^ 2 := by <proof>

theorem function_value_difference_rearrangement.{u_1} {F : Type u_1} [AddCommGroup F] [Module ℝ F] {f_k f_prev f_star : F} {γ_k : ℝ} :
  (f_k - f_star) - (1 - γ_k) • (f_prev - f_star) = γ_k • (f_k - f_star) + (1 - γ_k) • (f_k - f_prev) := by <proof>

theorem nesterov_convexity_application.{u_1} {E : Type u_1} [inst : NormedAddCommGroup E] [inst_1 : InnerProductSpace ℝ E] [inst_2 : CompleteSpace E] {f : E → ℝ} {f' : E → E} {x_star x₀ : E} {γ : ℕ+ → ℝ} {alg : Nesterov f f' γ x₀} (h_convex : ConvexOn ℝ univ f) (h_descent : ∀ (k : ℕ+) (x' : E), f (alg.x k) - f x' ≤ ↑alg.l * inner (alg.x k - alg.y k) (x' - alg.x k) + ↑alg.l / 2 * ‖alg.x k - alg.y k‖ ^ 2) (k : ℕ+) :
  f (alg.x k) - f x_star - (1 - γ k) * (f (alg.x (k - 1)) - f x_star) ≤
    ↑alg.l * inner (alg.x k - alg.y k) ((1 - γ k) • alg.x (k - 1) + (γ k) • x_star - alg.x k) + ↑alg.l / 2 * ‖alg.x k - alg.y k‖ ^ 2 := by <proof>

theorem integral_tilted_mean_formula.{u} {Ω : Type u} {m : MeasurableSpace Ω} {μ : Measure Ω}
  [IsFiniteMeasure μ] [NeZero μ] {a b : ℝ} {X : Ω → ℝ} (hX : AEMeasurable X μ)
  (h_bdd : ∀ᵐ ω ∂μ, X ω ∈ Set.Icc a b) :
  (fun t => ∫ ω, X ω ∂(μ.tilted (fun ω => t * X ω))) =
  (fun t => (∫ ω, Real.exp (t * X ω) * X ω ∂μ) / (∫ ω, Real.exp (t * X ω) ∂μ)) := by <proof>

theorem integral_tilted_second_moment_formula.{u} {Ω : Type u} {m : MeasurableSpace Ω} {μ : Measure Ω}
  [IsFiniteMeasure μ] [NeZero μ] {a b : ℝ} {X : Ω → ℝ} (hX : AEMeasurable X μ)
  (h_bdd : ∀ᵐ ω ∂μ, X ω ∈ Set.Icc a b) :
  (fun t => ∫ ω, (X ω) ^ 2 ∂(μ.tilted (fun ω => t * X ω))) =
  (fun t => (∫ ω, Real.exp (t * X ω) * (X ω) ^ 2 ∂μ) / (∫ ω, Real.exp (t * X ω) ∂μ)) := by <proof>

theorem inequality_for_strong_convex_lipschitz_gradient.{u_1} {E : Type u_1} [NormedAddCommGroup E] [InnerProductSpace ℝ E] {f : E → ℝ} {grad_f : E → E} {m : ℝ} {L : NNReal} {x x_minimizer : E} (h_strong_convex : StrongConvexOn Set.univ m f) (h_lipschitz_grad : LipschitzWith L grad_f) (h_grad_minimizer_zero : grad_f x_minimizer = 0) (hm_pos : m > 0) (hL_pos : 0 < (L : ℝ)) :
  inner (x - x_minimizer) (grad_f x) ≥
    m * (L : ℝ) / (m + L) * ‖x - x_minimizer‖ ^ 2 +
      1 / (m + L) * ‖grad_f x‖ ^ 2 := by <proof>

theorem gradient_step_algebraic_identity.{u_1} {E : Type u_1} [NormedAddCommGroup E] [InnerProductSpace ℝ E] {f : E → ℝ} {grad_f : E → E} {L : NNReal} {step_size : ℝ} (x : E) :
  f x + ⟪grad_f x, x - step_size • grad_f x - x⟫_ℝ + ↑L / 2 * ‖x - step_size • grad_f x - x‖ ^ 2 =
  f x + ((L : ℝ) / 2 * step_size * step_size - step_size) * ‖grad_f x‖ ^ 2 := by <proof>

theorem subgradient_bound_on_f_diff {E : Type*} [NormedAddCommGroup E] [InnerProductSpace ℝ E] {f : E → ℝ} {G : NNReal} (x y : E) (subgrad_at_x : E) (h_norm_bound : ‖subgrad_at_x‖ ≤ ↑G) (h_subgradient_def : f x - f y ≤ ⟪subgrad_at_x, x - y⟫_ℝ) :
  f x - f y ≤ ↑G * dist x y := by <proof>

theorem proximal_descent_lemma_relaxation.{u_1} {E : Type u_1} [NormedAddCommGroup E] [InnerProductSpace ℝ E] {f : E → ℝ} {f_grad_y : E} {x y : E} {L t_k : ℝ} (h_step_choice : L * t_k ≤ 1) :
  f y + inner f_grad_y (x - y) + L / 2 * ‖x - y‖ ^ 2 ≤ f y + inner f_grad_y (x - y) + 1 / (2 * t_k) * ‖x - y‖ ^ 2 := by <proof>
\end{minted}

\section{Proof Examples}
\label{app:proof_examples}

We show the proof examples using the lemma library in each domain.

\subsection{Inequality}

\begin{minted}[
    fontsize=\scriptsize,
    breaklines,
    breakanywhere,
    frame=lines,
    framesep=2mm,
    tabsize=2,
    samepage=false  % Allow page breaks
]{lean}
import Mathlib
import Aesop

set_option maxHeartbeats 0

open BigOperators Real Nat Topology Rat

-- Used lemmas from the library

theorem titu_lemma_three_reals
    {a b c x y z : ℝ}
    (ha : 0 < a) (hb : 0 < b) (hc : 0 < c) :
    (x + y + z) ^ 2 / (a + b + c) ≤ x ^ 2 / a + y ^ 2 / b + z ^ 2 / c := by <proof>

-- Main theorem

theorem MO_P8 {a b c : ℝ} (ha : a > 0) (hb : b > 0) (hc : c > 0) :
    1 ≤ a / Real.sqrt (a ^ 2 + 8 * b * c) +
        b / Real.sqrt (b ^ 2 + 8 * a * c) +
        c / Real.sqrt (c ^ 2 + 8 * a * b) := by
  let d1 := Real.sqrt (a ^ 2 + 8 * b * c)
  let d2 := Real.sqrt (b ^ 2 + 8 * a * c)
  let d3 := Real.sqrt (c ^ 2 + 8 * a * b)

  have hd1 : 0 < d1 := by
    dsimp [d1]
    apply Real.sqrt_pos.mpr
    positivity
  have hd2 : 0 < d2 := by
    dsimp [d2]
    apply Real.sqrt_pos.mpr
    positivity
  have hd3 : 0 < d3 := by
    dsimp [d3]
    apply Real.sqrt_pos.mpr
    positivity

  have hsum_pos : 0 < a + b + c := by
    linarith

  have h_titu :
      (a + b + c) ^ 2 / (a * d1 + b * d2 + c * d3) ≤
        a / d1 + b / d2 + c / d3 := by
    have h :=
      titu_lemma_three_reals
        (a := a * d1) (b := b * d2) (c := c * d3)
        (x := a) (y := b) (z := c)
        (by positivity) (by positivity) (by positivity)
    have h1 : a ^ 2 / (a * d1) = a / d1 := by
      field_simp [ha.ne', hd1.ne']
      ring
    have h2 : b ^ 2 / (b * d2) = b / d2 := by
      field_simp [hb.ne', hd2.ne']
      ring
    have h3 : c ^ 2 / (c * d3) = c / d3 := by
      field_simp [hc.ne', hd3.ne']
      ring
    simpa [h1, h2, h3] using h

  have h_cube :
      a ^ 3 + b ^ 3 + c ^ 3 + 24 * a * b * c ≤ (a + b + c) ^ 3 := by
    nlinarith [sq_nonneg (a - b), sq_nonneg (b - c), sq_nonneg (c - a)]

  have h_cs0 :=
    titu_lemma_three_reals
      (a := a) (b := b) (c := c)
      (x := a * d1) (y := b * d2) (z := c * d3)
      ha hb hc

  have h_sqdiv1 : (a * d1) ^ 2 / a = a * d1 ^ 2 := by
    field_simp [ha.ne']
    ring
  have h_sqdiv2 : (b * d2) ^ 2 / b = b * d2 ^ 2 := by
    field_simp [hb.ne']
    ring
  have h_sqdiv3 : (c * d3) ^ 2 / c = c * d3 ^ 2 := by
    field_simp [hc.ne']
    ring

  have h_cs :
      (a * d1 + b * d2 + c * d3) ^ 2 ≤
        (a + b + c) * (a * d1 ^ 2 + b * d2 ^ 2 + c * d3 ^ 2) := by
    have h_cs1 :
        (a * d1 + b * d2 + c * d3) ^ 2 / (a + b + c) ≤
          a * d1 ^ 2 + b * d2 ^ 2 + c * d3 ^ 2 := by
      simpa [h_sqdiv1, h_sqdiv2, h_sqdiv3, add_comm, add_left_comm, add_assoc] using h_cs0
    have := h_cs1
    rw [div_le_iff hsum_pos] at this
    simpa [mul_comm, mul_left_comm, mul_assoc] using this

  have h_expand :
      a * d1 ^ 2 + b * d2 ^ 2 + c * d3 ^ 2 =
        a ^ 3 + b ^ 3 + c ^ 3 + 24 * a * b * c := by
    dsimp [d1, d2, d3]
    rw [Real.sq_sqrt (by positivity), Real.sq_sqrt (by positivity), Real.sq_sqrt (by positivity)]
    ring

  have h_upper_sq :
      (a * d1 + b * d2 + c * d3) ^ 2 ≤ ((a + b + c) ^ 2) ^ 2 := by
    rw [h_expand] at h_cs
    have hmul :
        (a + b + c) * (a ^ 3 + b ^ 3 + c ^ 3 + 24 * a * b * c) ≤
          (a + b + c) * (a + b + c) ^ 3 := by
      gcongr
    have := le_trans h_cs hmul
    simpa [pow_succ, mul_assoc] using this

  have h_upper :
      a * d1 + b * d2 + c * d3 ≤ (a + b + c) ^ 2 := by
    have hnonneg : 0 ≤ a * d1 + b * d2 + c * d3 := by positivity
    nlinarith [h_upper_sq]

  have hden_pos : 0 < a * d1 + b * d2 + c * d3 := by positivity

  have hfrac :
      1 ≤ (a + b + c) ^ 2 / (a * d1 + b * d2 + c * d3) := by
    rw [le_div_iff hden_pos]
    simpa using h_upper

  have hmain :
      1 ≤ a / d1 + b / d2 + c / d3 := by
    exact le_trans hfrac h_titu

  simpa [d1, d2, d3] using hmain
\end{minted}

\subsection{Number Theory}

\begin{minted}[
    fontsize=\scriptsize,
    breaklines,
    breakanywhere,
    frame=lines,
    framesep=2mm,
    tabsize=2,
    samepage=false  % Allow page breaks
]{lean}
import Mathlib
import Aesop

set_option maxHeartbeats 0

open BigOperators Real Nat Topology Rat

-- Used lemmas from the library

theorem dirichlet_theorem_on_arithmetic_progressions (a n : ℕ) (h : Nat.Coprime a n) :
  Set.Infinite {p : ℕ | Nat.Prime p ∧ p ≡ a [MOD n]} := by <proof>

-- Main theorem

theorem infinitely_many_primes_10k_plus_9 :
  ∃ (S : Set ℕ), Set.Infinite S ∧ ∀ p ∈ S, Nat.Prime p ∧ p ≡ 9 [MOD 10] := by
  let S : Set ℕ := {p : ℕ | Nat.Prime p ∧ p ≡ 9 [MOD 10]}
  refine ⟨S, ?_, ?_⟩
  · have hcop : Nat.Coprime 9 10 := by
      norm_num
    simpa [S] using dirichlet_theorem_on_arithmetic_progressions 9 10 hcop
  · intro p hp
    exact hp
\end{minted}

\begin{minted}[
    fontsize=\scriptsize,
    breaklines,
    breakanywhere,
    frame=lines,
    framesep=2mm,
    tabsize=2,
    samepage=false  % Allow page breaks
]{lean}
import Mathlib
import Aesop

set_option maxHeartbeats 0

open BigOperators Real Nat Topology Rat

-- Used lemmas from the library

theorem coprime_iff_forall_prime_not_dvd (a n : ℤ) (hn' : n.natAbs ≠ 0) :
  Nat.gcd a.natAbs n.natAbs = 1 ↔ ∀ p : ℕ, p.Prime → (p : ℤ) ∣ n → ¬ ((p : ℤ) ∣ a) := by <proof>

-- Main theorem

theorem integer_representation_theorem (l n : ℤ) (hl : 1 ≤ l ∧ l ≤ n) (hn : n ≥ 3 ∧ n % 2 = 1) :
  ∃ (a b : ℤ), a < n ∧ b < n ∧
  Nat.gcd (a.natAbs) n.natAbs = 1 ∧
  Nat.gcd (b.natAbs) n.natAbs = 1 ∧
  (l = a + b ∨ l = a - b) := by
  obtain ⟨hl1, hl2⟩ := hl
  obtain ⟨hn3, hnodd⟩ := hn
  have hnpos : n > 0 := by
    linarith
  have hnne : n ≠ 0 := by
    linarith
  have hnabs_ne : n.natAbs ≠ 0 := by
    exact Int.natAbs_ne_zero.mpr hnne
  have hchoices :
      ∀ p : ℕ, p.Prime → (p : ℤ) ∣ n →
        ∃ c : ℤ, ¬ ((p : ℤ) ∣ c) ∧ ¬ ((p : ℤ) ∣ (l - c)) := by
    intro p hp hpn
    have hp_ge_3 : p ≥ 3 := by
      exact prime_divisor_of_odd_integer_ge_three n p hnodd hp hpn
    exact exists_residue_avoiding_two_values p l hp hp_ge_3
  obtain ⟨a, ha1, ha_le_n, ha_cop, hla_cop⟩ :=
    exists_int_by_crt_from_local_choices n l hnpos hchoices
  have hgcd_a : Nat.gcd a.natAbs n.natAbs = 1 := by
    exact (coprime_iff_forall_prime_not_dvd a n hnabs_ne).mpr ha_cop
  have hgcd_b : Nat.gcd (l - a).natAbs n.natAbs = 1 := by
    exact (coprime_iff_forall_prime_not_dvd (l - a) n hnabs_ne).mpr hla_cop
  refine ⟨a, l - a, ?_, ?_, hgcd_a, hgcd_b, ?_⟩
  · have hane : a ≠ n := by
      apply int_ne_of_coprime a n
      · linarith
      · linarith
      · exact hgcd_a
    exact int_lt_of_le_ne a n ha_le_n hane
  · have hbound : l - a ≤ n - 1 := by
      exact int_bound_on_difference l a n hl2 ha1
    linarith
  · left
    ring
\end{minted}

\subsection{Combinatorics}

\begin{minted}[
    fontsize=\scriptsize,
    breaklines,
    breakanywhere,
    frame=lines,
    framesep=2mm,
    tabsize=2,
    samepage=false  % Allow page breaks
]{lean}
import Mathlib
import Aesop

open BigOperators Classical ENNReal Equiv EuclideanGeometry Filter Finset Fintype Function Lex List MeasureTheory Nat ProbabilityTheory Real SimpleGraph Real Nat Topology Rat

set_option maxHeartbeats 0

-- Used lemmas from the library

theorem card_positive_solutions_sum_eq_choose {n k : ℕ} (h_k_pos : k > 0) :
  (Finset.univ.filter (fun f : Fin k → Fin (n + 1) => (∀ i, (f i : ℕ) > 0) ∧ ∑ i, (f i : ℕ) = n)).card = Nat.choose (n - 1) (k - 1) := by <proof>

-- Main theorem

theorem hackmath_9 (sols : Finset (Fin 6 → ℕ))
    (h_sols : ∀ f, f ∈ sols ↔ ((∀ i, f i > 0) ∧ (∑ i, f i = 10))) :
    sols.card = ((126) : ℕ ) := by
  classical
  let t : Finset (Fin 6 → Fin 11) :=
    Finset.univ.filter (fun f : Fin 6 → Fin 11 => (∀ i, (f i : ℕ) > 0) ∧ ∑ i, (f i : ℕ) = 10)

  have hcard_t : t.card = 126 := by
    rw [card_positive_solutions_sum_eq_choose (n := 10) (k := 6)]
    · native_decide
    · native_decide

  have hmem_range : ∀ f : Fin 6 → ℕ, ((∀ i, f i > 0) ∧ ∑ i, f i = 10) → ∀ i, f i < 11 := by
    intro f hf i
    obtain ⟨_, hsum⟩ := hf
    have hsingle : f i ≤ ∑ j, f j := by
      simpa using
        (Finset.single_le_sum (fun j _ => Nat.zero_le (f j))
          (by simp : i ∈ (Finset.univ : Finset (Fin 6))))
    rw [hsum] at hsingle
    omega

  have hcard_eq : sols.card = t.card := by
    refine Finset.card_bij
      (fun f hf i => ⟨f i, hmem_range f ((h_sols f).1 hf) i⟩)
      ?_ ?_ ?_
    · intro f hf
      simp only [t, Finset.mem_filter, Finset.mem_univ, true_and]
      simpa using ((h_sols f).1 hf)
    · intro f hf f' hf' hEq
      funext i
      have hi := congrArg (fun g : Fin 6 → Fin 11 => ((g i : Fin 11) : ℕ)) hEq
      simpa using hi
    · intro g hg
      refine ⟨(fun i => (g i : ℕ)), ?_, ?_⟩
      · have hg' : (∀ i, (g i : ℕ) > 0) ∧ ∑ i, (g i : ℕ) = 10 := by
          simpa only [t, Finset.mem_filter, Finset.mem_univ, true_and] using hg
        exact (h_sols (fun i => (g i : ℕ))).2 hg'
      · funext i
        rfl

  calc
    sols.card = t.card := hcard_eq
    _ = 126 := hcard_t
\end{minted}

\begin{minted}[
    fontsize=\scriptsize,
    breaklines,
    breakanywhere,
    frame=lines,
    framesep=2mm,
    tabsize=2,
    samepage=false  % Allow page breaks
]{lean}
import Mathlib
import Aesop

open BigOperators Classical ENNReal Equiv EuclideanGeometry Filter Finset Fintype Function Lex List MeasureTheory Nat ProbabilityTheory Real SimpleGraph Real Nat Topology Rat

set_option maxHeartbeats 0

-- Used lemmas from the library

theorem Finset.card_eq_of_mem_powersetCard {α : Type*} {X : Finset α} {r : ℕ} {s : Finset α} :
    s ∈ X.powersetCard r → s.card = r := by <proof>

theorem Finset.lex_le_of_complement_of_lex_le {α : Type*} [Fintype α] [LinearOrder α] {s t : Finset α} :
    s.card = t.card → List.Lex (· ≤ ·) (s.sort (· ≤ ·)) (t.sort (· ≤ ·)) → List.Lex (· ≤ ·) (tᶜ.sort (· ≤ ·)) (sᶜ.sort (· ≤ ·)) := by <proof>

-- Main theorem

theorem brualdi_ch4_35 (r n M : ℕ) (hM : M = ((@Finset.univ (Fin n)).powersetCard r).card)
    (A : Fin M → (Finset.powersetCard r (@Finset.univ (Fin M) _))) :
    ∀ i j, (List.Lex (fun x1 x2 : Fin M => x1 ≤ x2)
    (Finset.sort (· ≤ ·) (A i)) (Finset.sort (· ≤ ·) (A j))) →
    (List.Lex (fun x1 x2 : Fin M => x1 ≤ x2)
    (Finset.sort (· ≤ ·) (A j)ᶜ) (Finset.sort (· ≤ ·) (A i)ᶜ)) := by
  classical
  intro i j hlex
  have hi : ((A i : Finset (Fin M))).card = r := by
    exact Finset.card_eq_of_mem_powersetCard (A i).2
  have hj : ((A j : Finset (Fin M))).card = r := by
    exact Finset.card_eq_of_mem_powersetCard (A j).2
  have hcards : ((A i : Finset (Fin M))).card = ((A j : Finset (Fin M))).card := by
    rw [hi, hj]
  convert
    (Finset.lex_le_of_complement_of_lex_le
      (s := (A i : Finset (Fin M)))
      (t := (A j : Finset (Fin M)))
      hcards
      hlex)
\end{minted}

\subsection{Geometry}

\begin{minted}[
    fontsize=\scriptsize,
    breaklines,
    breakanywhere,
    frame=lines,
    framesep=2mm,
    tabsize=2,
    samepage=false  % Allow page breaks
]{lean}
import Mathlib
import SystemE
import LeanGeo
import Geo
namespace LeanGeo

set_option maxHeartbeats 0

-- Used lemmas from the library

theorem formTriangle_imp_triangle (A B C : Point) (l_ab l_bc l_ca : Line) :
  formTriangle A B C l_ab l_bc l_ca →
  Triangle A B C := by <proof>

theorem circumcenter_of_incenter_and_two_vertices_on_circumcircle (A B C I O₁ : Point) (Γ : Circle) :
  Triangle A B C →
  Incentre I A B C →
  Circumcentre O₁ I A B →
  Circumcircle Γ A B C →
  O₁.onCircle Γ := by <proof>

theorem dist_center_point_on_circle_eq_radius (O P : Point) (Γ : Circle) :
  O.isCentre Γ →
  P.onCircle Γ →
  |(O─P)| = rad Γ := by <proof>

-- Main theorem

theorem Problem_4_42_USAMO_1988_4 :
  ∀ (A B C I O O1 O2 O3 : Point) (AB BC CA : Line),
    formTriangle A B C AB BC CA →
    Incentre I A B C →
    Circumcentre O A B C →
    Circumcentre O1 I A B →
    Circumcentre O2 I B C →
    Circumcentre O3 I C A →
    |(O─O1)| = |(O─O2)| ∧ |(O─O2)| = |(O─O3)| := by
  euclid_intros
  have h_tri : Triangle A B C := by
    euclid_apply formTriangle_imp_triangle A B C AB BC CA
    euclid_finish
  euclid_apply threePoints_existCircle A B C as Γ
  have h_circum_Γ : Circumcircle Γ A B C := by euclid_finish
  have h_O_is_center : O.isCentre Γ := by
    euclid_apply circumcentre_isCentre_circumcircle A B C O Γ
    euclid_finish
  have h_O1_on_Γ : O1.onCircle Γ := by
    euclid_apply circumcenter_of_incenter_and_two_vertices_on_circumcircle A B C I O1 Γ
    euclid_finish
  have h_O2_on_Γ : O2.onCircle Γ := by
    euclid_apply circumcenter_of_incenter_and_two_vertices_on_circumcircle B C A I O2 Γ
    euclid_finish
  have h_O3_on_Γ : O3.onCircle Γ := by
    euclid_apply circumcenter_of_incenter_and_two_vertices_on_circumcircle C A B I O3 Γ
    euclid_finish
  have h_dist1 : |(O─O1)| = rad Γ := by
    euclid_apply dist_center_point_on_circle_eq_radius O O1 Γ
    euclid_finish
  have h_dist2 : |(O─O2)| = rad Γ := by
    euclid_apply dist_center_point_on_circle_eq_radius O O2 Γ
    euclid_finish
  have h_dist3 : |(O─O3)| = rad Γ := by
    euclid_apply dist_center_point_on_circle_eq_radius O O3 Γ
    euclid_finish
  euclid_finish
\end{minted}

\begin{minted}[
    fontsize=\scriptsize,
    breaklines,
    breakanywhere,
    frame=lines,
    framesep=2mm,
    tabsize=2,
    samepage=false  % Allow page breaks
]{lean}
import Mathlib
import SystemE
import LeanGeo
import Geo
namespace LeanGeo

set_option maxHeartbeats 0

-- Used lemmas from the library

theorem point_inside_circle_on_secant_is_between (A P I : Point) (L : Line) (Ω : Circle)
  (h_I_inside : I.insideCircle Ω)
  (h_A_on : A.onCircle Ω)
  (h_P_on : P.onCircle Ω)
  (h_coll : A.onLine L ∧ I.onLine L ∧ P.onLine L)
  (h_ne : P ≠ A) :
  between A I P := by <proof>

theorem power_of_point_interior_formulation (A O I P : Point) (Ω : Circle)
  (h_is_centre : O.isCentre Ω)
  (h_A_on : A.onCircle Ω)
  (h_P_on : P.onCircle Ω)
  (h_between : between A I P) :
  |(O─A)| * |(O─A)| - |(O─I)| * |(O─I)| = |(A─I)| * |(I─P)| := by <proof>

theorem power_of_incenter_equals_2Rr (A B C O I D P : Point) (BC : Line) (Ω : Circle)
  (h_circumcenter : Circumcentre O A B C)
  (h_incenter : Incentre I A B C)
  (h_foot : Foot I D BC)
  (h_is_centre : O.isCentre Ω)
  (h_circumcircle : A.onCircle Ω ∧ B.onCircle Ω ∧ C.onCircle Ω)
  (h_P_on : P.onCircle Ω)
  (h_coll_AIP : Coll A I P)
  (h_ne : P ≠ A) :
  |(A─I)| * |(I─P)| = 2 * |(O─A)| * |(I─D)| := by <proof>

-- Main theorem

theorem Euler's_Theorem :
  ∀ (A B C O I D : Point) (AB BC CA : Line),
    formTriangle A B C AB BC CA ∧
    Circumcentre O A B C ∧
    Incentre I A B C ∧
    Foot I D BC →
    |(O─I)| * |(O─I)| = |(O─A)| * (|(O─A)| - 2 * |(I─D)|) := by
  euclid_intros
  have h_tri : Triangle A B C := by
    euclid_apply formTriangle_implies_triangle A B C AB BC CA
    euclid_finish
  euclid_apply threePoints_existCircle A B C as Ω
  have h_O_is_centre : O.isCentre Ω := by
    euclid_apply circumcentre_isCentre_circumcircle A B C O Ω
    euclid_finish
  have h_A_ne_I : A ≠ I := by
    euclid_apply incenter_is_distinct_from_vertex A B C I
    euclid_finish
  euclid_apply line_from_points A I as AI_line
  have h_I_inside : I.insideCircle Ω := by
    euclid_apply incenter_inside_circumcircle A B C I O Ω
    euclid_finish
  have h_intersects : AI_line.intersectsCircle Ω := by
    euclid_finish
  euclid_apply intersections_circle_line Ω AI_line as (P, Q)
  have h_A_is_one : (P = A ∧ Q ≠ A) ∨ (Q = A ∧ P ≠ A) := by
    euclid_apply line_circle_intersections_are_unique A P Q AI_line Ω
    euclid_finish
  cases h_A_is_one
  case inl h_P_is_A =>
    have h_between_AIQ : between A I Q := by
      euclid_apply point_inside_circle_on_secant_is_between A Q I AI_line Ω
      euclid_finish
    have h_pow_I : |(O─A)| * |(O─A)| - |(O─I)| * |(O─I)| = |(A─I)| * |(I─Q)| := by
      euclid_apply power_of_point_interior_formulation A O I Q Ω
      euclid_finish
    have h_2Rr : |(A─I)| * |(I─Q)| = 2 * |(O─A)| * |(I─D)| := by
      euclid_apply power_of_incenter_equals_2Rr A B C O I D Q BC Ω
      euclid_finish
    euclid_assert |(O─A)| * |(O─A)| - |(O─I)| * |(O─I)| = 2 * |(O─A)| * |(I─D)|
    euclid_finish
  case inr h_Q_is_A =>
    have h_between_AIP : between A I P := by
      euclid_apply point_inside_circle_on_secant_is_between A P I AI_line Ω
      euclid_finish
    have h_pow_I : |(O─A)| * |(O─A)| - |(O─I)| * |(O─I)| = |(A─I)| * |(I─P)| := by
      euclid_apply power_of_point_interior_formulation A O I P Ω
      euclid_finish
    have h_2Rr : |(A─I)| * |(I─P)| = 2 * |(O─A)| * |(I─D)| := by
      euclid_apply power_of_incenter_equals_2Rr A B C O I D P BC Ω
      euclid_finish
    euclid_assert |(O─A)| * |(O─A)| - |(O─I)| * |(O─I)| = 2 * |(O─A)| * |(I─D)|
    euclid_finish
\end{minted}

\subsection{Machine Learning Theory}

\begin{minted}[
    fontsize=\scriptsize,
    breaklines,
    breakanywhere,
    frame=lines,
    framesep=2mm,
    tabsize=2,
    samepage=false  % Allow page breaks
]{lean}
import Mathlib
import FoML
import Optlib
import ML

open scoped ENNReal MeasureTheory NNReal Pointwise ProbabilityTheory Topology
open Asymptotics BCD BigOperators Bornology ContinuousLinearMap ENNReal Filter Finset Function InnerProduct InnerProductSpace LinearMap Matrix MeasureTheory Metric NNReal PNat ProbabilityTheory Real Set Topology

-- Used lemmas from the library

theorem inequality_for_strong_convex_lipschitz_gradient.{u_1} {E : Type u_1} [NormedAddCommGroup E] [InnerProductSpace ℝ E] {f : E → ℝ} {grad_f : E → E} {m : ℝ} {L : NNReal} {x x_minimizer : E} (h_strong_convex : StrongConvexOn Set.univ m f) (h_lipschitz_grad : LipschitzWith L grad_f) (h_grad_minimizer_zero : grad_f x_minimizer = 0) (hm_pos : m > 0) (hL_pos : 0 < (L : ℝ)) :
  inner (x - x_minimizer) (grad_f x) ≥
    m * (L : ℝ) / (m + L) * ‖x - x_minimizer‖ ^ 2 +
      1 / (m + L) * ‖grad_f x‖ ^ 2 := by <proof>

-- Main theorem

theorem GD_0.{u_1} {E : Type u_1} [inst : NormedAddCommGroup E] [inst_1 : InnerProductSpace ℝ E]
  [inst_2 : CompleteSpace E] {f : E → ℝ} {m : ℝ} {f' : E → E} {xm x₀ : E} {alg : Gradient_Descent_fix_stepsize f f' x₀}
  (hm : m > 0) (min : IsMinOn f univ xm)
  (step₂ : Gradient_Descent_fix_stepsize.a f f' x₀ ≤ 2 / (m + ↑(Gradient_Descent_fix_stepsize.l f f' x₀)))
  (hsc : StrongConvexOn univ m f) (this_1 : LipschitzWith (Gradient_Descent_fix_stepsize.l f f' x₀) f')
  (this : ↑(Gradient_Descent_fix_stepsize.l f f' x₀) > 0) (k : ℕ)
  (eq1 : f' (Gradient_Descent_fix_stepsize.x f f' x₀ k) = f' (Gradient_Descent_fix_stepsize.x f f' x₀ k) - f' xm) :
  inner (Gradient_Descent_fix_stepsize.x f f' x₀ k - xm) (f' (Gradient_Descent_fix_stepsize.x f f' x₀ k)) ≥
    m * ↑(Gradient_Descent_fix_stepsize.l f f' x₀) / (m + ↑(Gradient_Descent_fix_stepsize.l f f' x₀)) *
        ‖Gradient_Descent_fix_stepsize.x f f' x₀ k - xm‖ ^ 2 +
      1 / (m + ↑(Gradient_Descent_fix_stepsize.l f f' x₀)) * ‖f' (Gradient_Descent_fix_stepsize.x f f' x₀ k)‖ ^ 2 := by
  have h_grad_minimizer_zero : f' xm = 0 :=
    (sub_eq_self.mp eq1.symm)
  exact
    inequality_for_strong_convex_lipschitz_gradient hsc this_1 h_grad_minimizer_zero hm this
\end{minted}

\begin{minted}[
    fontsize=\scriptsize,
    breaklines,
    breakanywhere,
    frame=lines,
    framesep=2mm,
    tabsize=2,
    samepage=false  % Allow page breaks
]{lean}
import Mathlib
import FoML
import Optlib
import ML

open scoped ENNReal MeasureTheory NNReal Pointwise ProbabilityTheory Topology
open Asymptotics BCD BigOperators Bornology ContinuousLinearMap ENNReal Filter Finset Function InnerProduct InnerProductSpace LinearMap Matrix MeasureTheory Metric NNReal PNat ProbabilityTheory Real Set Topology

-- Used lemmas from the library

theorem quadratic_form_difference_identity {m n : ℕ+} (A : Matrix (Fin m) (Fin n) ℝ) (x y : EuclideanSpace ℝ (Fin n)) :
  A *ᵥ y ⬝ᵥ A *ᵥ y - A *ᵥ x ⬝ᵥ A *ᵥ x - 2 * ((Aᵀ * A) *ᵥ x ⬝ᵥ (y - x)) = ‖A *ᵥ y - A *ᵥ x‖ ^ 2 := by <proof>

-- Main theorem

theorem LASSO_1 {n m : ℕ+} {A : Matrix (Fin ↑m) (Fin ↑n) ℝ} (hA : ¬A = 0)
  (x : EuclideanSpace ℝ (Fin ↑n)) (ε : ℝ) (εpos : ε > 0) :
  let normA := ‖(toEuclideanLin ≪≫ₗ LinearMap.toContinuousLinearMap) A‖;
  (∀ (x : EuclideanSpace ℝ (Fin ↑n)), ‖A *ᵥ x‖ ≤ normA * ‖x‖) →
    0 < normA →
      ∀ (y : EuclideanSpace ℝ (Fin ↑n)),
        ‖x - y‖ ≤ ε / normA ^ 2 →
          ((fun x_1 => y x_1 - x x_1) ⬝ᵥ fun x_1 => ((Aᵀ * A) *ᵥ x) x_1) = (Aᵀ * A) *ᵥ x ⬝ᵥ (y - x) →
            |A *ᵥ y ⬝ᵥ A *ᵥ y - A *ᵥ x ⬝ᵥ A *ᵥ x - 2 * (fun x_1 => y x_1 - x x_1) ⬝ᵥ (Aᵀ * A) *ᵥ x| ≤ ε * ‖x - y‖ →
              ‖A *ᵥ y - A *ᵥ x‖ ^ 2 ≤ ε * ‖x - y‖ := by
  intros normA h_norm_bound h_normA_pos y h_dist h_dot_product_rewrite h_abs
  have h_rw : (fun x_1 => y x_1 - x x_1) ⬝ᵥ (Aᵀ * A) *ᵥ x = (Aᵀ * A) *ᵥ x ⬝ᵥ (y - x) := by
    trans ((fun x_1 => y x_1 - x x_1) ⬝ᵥ fun x_1 => ((Aᵀ * A) *ᵥ x) x_1)
    · rfl
    · exact h_dot_product_rewrite
  rw [h_rw] at h_abs
  rw [quadratic_form_difference_identity] at h_abs
  rw [abs_of_nonneg (sq_nonneg _)] at h_abs
  exact h_abs
\end{minted}

\section{Prompts for LLM-Based Components}
\label{app:prompts}

We list the relevant prompt templates used in our agentic framework.

\subsection{Whole Proof Generation}

\begin{tcolorbox}[colback=gray!10, colframe=gray!60, coltitle=black,
                  title=Prompt for Whole Proof Generation, fonttitle=\bfseries]
\begin{Verbatim}[fontsize=\scriptsize, breaklines=true]
You are a Lean 4 expert who is trying to help write a proof in Lean 4. Think step-by-step to complete the following Lean 4 proof.

Problem Statement:
{problem}

Instructions:
Same proof level = same indentation: All tactics at the same logical level must use identical indentation
Consistent characters: Use either tabs OR spaces consistently (don't mix)
Proper nesting: Indent sub-proofs one level deeper than their parent
Do NOT include any imports or open statements.
Use proper Lean 4 syntax and conventions. Ensure the proof sketch is enclosed in triple backticks ```lean```. 
Only include a single Lean 4 code block, corresponding to the proof along with the theorem statement.
Do NOT use sorry.
Do NOT change anything in the original theorem statement.

Here is a list of useful and proven theorems that might be helpful to solve the problem.
You can use them without proving them.
{useful_theorem_section}

\end{Verbatim}
\end{tcolorbox}

\begin{tcolorbox}[colback=gray!10, colframe=gray!60, coltitle=black,
                  title=Prompt for Error Correction, fonttitle=\bfseries]
\begin{Verbatim}[fontsize=\scriptsize, breaklines=true]
You are a Lean 4 expert who is trying to help write a proof in Lean 4. The following Lean 4 code has compilation errors. Please fix the errors while maintaining the mathematical meaning. 

{error_message}

Instructions:
Analyze what the theorem is trying to prove. Then, analyze why the error is happening, step-by-step. Add a brief explanation.
Then, provide a corrected version of the Lean 4 code that addresses these specific errors.
Do NOT include any other Lean code blocks except for the proof.
Do NOT use sorry.
Do NOT include any imports or open statements.
Do NOT change anything in the original theorem statement.

Here is a list of useful and proven theorems that might be helpful to solve the problem.
You can use them without proving them.
{useful_theorems_section}

\end{Verbatim}
\end{tcolorbox}

\subsection{Theorem Decomposition}

\begin{tcolorbox}[colback=gray!10, colframe=gray!60, coltitle=black,
                  title=Prompt for Sketch Generation, fonttitle=\bfseries]
\begin{Verbatim}[fontsize=\scriptsize, breaklines=true]
You are a mathematical and Lean 4 expert whose goal is to solve problems with rigorous mathematical reasoning.

Problem Statement:
{problem}

Instructions:
First provide a natural language, step-by-step proof for the given problem. Start from the given premises and reason step-by-step to reach the conclusion.
Use the informal proof to write a proof sketch for the problem in Lean 4 following these 
Break complex reasoning into logical sub-goals using `have` statements. The subgoals should build up to prove the main theorem. Make sure to include all the steps and calculations from the given proof in the proof sketch. Each subgoal should ideally require applying just one key theorem or lemma, or a few tactic applications.

Do NOT create subgoals identical to any of the given hypotheses
Do NOT create subgoals that are more complex than the original problems. The subgoals should be SIMPLER than the given problem.
Do NOT skip over any steps. Do NOT make any mathematical leaps.

Rules:
Same proof level = same indentation: All tactics at the same logical level must use identical indentation
Consistent characters: Use either tabs OR spaces consistently (don't mix)
Proper nesting: Indent sub-proofs one level deeper than their parent
Do NOT nest `have` statements in each other. Use distinct sub-goals as much as possible. Ensure all sub goals are named. Do NOT create anonymous have statements.
 Do NOT include any imports or open statements in your code.
One line = One `have` subgoal. Do NOT split subgoals across different lines.
Use proper Lean 4 syntax and conventions. Ensure the proof sketch is enclosed in triple backticks ```lean```
Use `sorry` for all subgoal proofs - focus on structure, not implementation
**Do NOT use `sorry` for the main goal proof** - use your subgoals to prove it
NEVER use `sorry` IN the theorem statement itself
Ensure subgoals collectively provide everything needed for the main proof
Make the logical dependencies between subgoals explicit. Ensure that the subgoals are valid and provable in Lean 4.
Do NOT change anything in the original theorem statement.

Here is a list of useful and proven theorems that might be helpful to solve the problem.
You can use them without proving them.
{useful_theorems_section}

\end{Verbatim}
\end{tcolorbox}

\begin{tcolorbox}[colback=gray!10, colframe=gray!60, coltitle=black,
                  title=Prompt for Sketch Correction, fonttitle=\bfseries]
\begin{Verbatim}[fontsize=\scriptsize, breaklines=true]
You are a Lean 4 expert who is trying to help write a proof in Lean 4. The following Lean 4 code has compilation errors. Please fix the errors while maintaining the mathematical meaning. 

{error_message}

Instructions:
Analyze what the theorem is trying to prove. Then, analyze why the error is happening, step-by-step. Add a brief explanation.
Then, provide a corrected version of the Lean 4 code that addresses these specific errors.
Do NOT include any other Lean code blocks except for the proof. Do NOT include any imports or open statements.
Use sorry for the proof of all `have` statements.
Ensure there are no use of `sorry` statements outside of `have` statements. Do NOT use `sorry` while proving the main theorem.
Do NOT change anything in the original theorem statement.
Do NOT nest `have` statements in each other. Use distinct sub-goals as much as possible. Ensure all sub goals are named. Do NOT create anonymous have statements.

Here is a list of useful and proven theorems that might be helpful to solve the problem.
You can use them without proving them.
{useful_theorems_section}

\end{Verbatim}
\end{tcolorbox}

\begin{tcolorbox}[colback=gray!10, colframe=gray!60, coltitle=black,
                  title=Prompt for Theorem Decomposition, fonttitle=\bfseries]
\begin{Verbatim}[fontsize=\scriptsize, breaklines=true]
From this proof sketch, extract any missing proofs (specified with `sorry`) as independent subgoals (theorems).

Proof Sketch:
```lean4
{proof_sketch}
```

Instructions:
Use the same name as the have statements for the theorems.
Each subgoal should have the relevant context from the previous subgoals needed to simplify the proof as much as possible.
There should be as many extracted theorems as `sorry`s in the given theorem.
Do NOT include any imports or open statements. Do NOT add any definitions. ONLY include the theorem statement.
Use a separate Lean 4 ``lean`` block for each subgoal.
Use sorry for the proof. Do NOT prove any theorem.
Do NOT change the conclusion of the theorems from the extracted subgoals. Keep them AS IT IS.
Do NOT change the conclusions of the preceding theorems when presenting them as hypotheses for the next subgoals. Keep them AS IT IS.
Do NOT duplicate theorem names. Use distinct theorem names for the different theorems.
Make sure the names and types of the premises/arguments in the extracted theorems MATCH the subgoals from which they are extracted.

\end{Verbatim}
\end{tcolorbox}

\begin{tcolorbox}[colback=gray!10, colframe=gray!60, coltitle=black,
                  title=Prompt for Proof Assembly, fonttitle=\bfseries]
\begin{Verbatim}[fontsize=\scriptsize, breaklines=true]
You are a Lean 4 expert. Your goal is to write a proof in Lean 4, according to the given proof sketch, using the supplied theorems. Your answer should be a single Lean 4 block containing the completed proof for the given theorem.

Proof sketch:
```lean4
{proof_sketch}
```

Theorems:
```lean4
{theorems_string}
```

Instructions:
You can assume that the theorems are correct and use them directly in your proof.
Do NOT modify the given theorems.
Do NOT prove the given theorems. 
Do NOT modify the given proof sketch steps. Simply apply the given theorems to complete the missing `sorry` steps.
Do NOT use `sorry` in your proof.
Do NOT include any imports or definitions or open statements.
Do NOT re-define the given theorems in your response.
Do NOT write a proof for any subgoal from scratch. ALWAYS use the supplied theorems.

\end{Verbatim}
\end{tcolorbox}

\subsection{Lemma Evolving}

\begin{tcolorbox}[colback=gray!10, colframe=gray!60, coltitle=black,
                  title=Prompt for Lemma Annotation, fonttitle=\bfseries]
\begin{Verbatim}[fontsize=\scriptsize, breaklines=true]
You are a mathematician and an expert in Lean. Below is a subgoal in a problem that you are going to prove:

```lean4
{proof}
```

Generate a natural language description of the theorem.
First determine the sub-domain of the theorem.
Then determine its difficulty, whether it's trivial or not, e.g. equivalence transformation. Highlight the use of well-know lemma or their variants that can be used to prove it if it's non-trivial.
Then summarize to a brief description.
Here are some instructions of the final summarized description:

Instructions:
1. Keep the description short, within 1 to 5 sentences.
2. Introduce the name if it is some well-known theorem or one of its variants.
2. Enclose the description in one pair of <description> tags.

\end{Verbatim}
\end{tcolorbox}

\begin{tcolorbox}[colback=gray!10, colframe=gray!60, coltitle=black,
                  title=Prompt for Cluster-based Lemma Abstraction, fonttitle=\bfseries]
\begin{Verbatim}[fontsize=\scriptsize, breaklines=true]
You are a mathematician and an expert in Lean. Below are a set of theorems that you proved:

{theorem_section}

Please propose new general theorems that are closely related to the above theorems.
Propose only theorems in the {domain} domain.
The proposed theorems should be general and reusable, and can be used to prove the above theorems.
Below are some instructions that you should follow.

Instructions:
1. Do NOT include any imports or open statements. Do NOT add any definitions. ONLY include the theorem statement.
2. Use a separate Lean 4 ``lean`` block for each theorem.
3. Do NOT prove any theorem. Do not include ":= by sorry" but only the statement.
4. Do NOT duplicate theorem names. Use distinct theorem names for the different theorems.
5. Do Not use general types, align types and number of variables with the ones in the provided theorems.
6. MAKE SURE that the proposed theorems are all syntactically and logically correct.

\end{Verbatim}
\end{tcolorbox}

\end{document}